\documentclass[10pt,twocolumn,letterpaper]{article}

\usepackage[pagenumbers]{cvpr} % To force page numbers, e.g. for an arXiv version
\usepackage{url}            % simple URL typesetting
\usepackage{booktabs}       % professional-quality tables
\usepackage{amsfonts}       % blackboard math symbols
\usepackage{nicefrac}       % compact symbols for 1/2, etc.
\usepackage{microtype}      % microtypography
\usepackage{xcolor}         % colors

\usepackage{siunitx}

\usepackage{adjustbox}
\usepackage{diagbox}
\usepackage{multirow}
\usepackage{makecell}

\usepackage{times}
\usepackage{epsfig}
\usepackage{graphicx}
\usepackage{amsmath}
\usepackage{amssymb}
\usepackage{colortbl}
\usepackage{xspace}

\usepackage{ragged2e}
\usepackage{float}
\usepackage{bbding}
\usepackage{amssymb}  
\usepackage[accsupp]{axessibility}
\usepackage{bbding}

\definecolor{cvprblue}{rgb}{0.21,0.49,0.74}
\usepackage[pagebackref,breaklinks,colorlinks,allcolors=cvprblue]{hyperref}
\definecolor{verylightgrey}{rgb}{0.95, 0.95, 0.95}
\definecolor{cvprblue}{rgb}{0.21,0.49,0.74}

\def\paperID{13108} % *** Enter the Paper ID here
\def\confName{CVPR}
\def\confYear{2025}

\title{CrayonRobo: Object-Centric Prompt-Driven Vision-Language-Action \\ Model for 
Robotic Manipulation}

\author{
Xiaoqi Li\textsuperscript{\rm 1,2}, 
Lingyun Xu\textsuperscript{\rm 1},
Mingxu Zhang\textsuperscript{\rm 1},
Jiaming Liu\textsuperscript{\rm 1},
Yan Shen\textsuperscript{\rm 1,2},
Iaroslav Ponomarenko\textsuperscript{\rm 1,2},\\
Jiahui Xu\textsuperscript{\rm 1},
Liang Heng\textsuperscript{\rm 1,2},
Siyuan Huang\textsuperscript{\rm 1},
Shanghang Zhang\textsuperscript{\rm 1},
Hao Dong\textsuperscript{\rm 1,2}\thanks{Corresponding author. First author's email: xl3062@columbia.edu.}
\\
\textsuperscript{\rm 1}School of Computer Science, Peking University 
\textsuperscript{\rm 2} PKU-Agibot Lab
}

\begin{document}
\maketitle
\begin{abstract}
In robotic, task goals can be conveyed through various modalities, such as language, goal images, and goal videos. However, natural language can be ambiguous, while images or videos may offer overly detailed specifications. 
To tackle these challenges, we introduce CrayonRobo that leverages comprehensive multi-modal prompts that explicitly convey both low-level actions and high-level planning in a simple manner.
Specifically, for each key-frame in the task sequence, our method allows for manual or automatic generation of simple and expressive 2D visual prompts overlaid on RGB images. 
These prompts represent the required task goals, such as the end-effector pose and the desired movement direction after contact. 
We develop a training strategy that enables the model to interpret these visual-language prompts and predict the corresponding contact poses and movement directions in SE(3) space.
Furthermore, by sequentially executing all key-frame steps, the model can complete long-horizon tasks. This approach not only helps the model explicitly understand the task objectives but also enhances its robustness on unseen tasks by providing easily interpretable prompts.
We evaluate our method in both simulated and real-world environments, demonstrating its robust manipulation capabilities.
\end{abstract}    
\section{Introduction}
\label{sec:intro}

As thoughtful helpers for humans, it is crucial for robots to understand and successfully execute their assigned tasks. Various approaches exist to convey the goals that robots should achieve, such as language descriptions, goal images, or goal videos. Language instructions ~\citep{liang2023code,ahn2022can,nair2022learning,lynch2020language,huang2023voxposer,shridhar2022cliport,shridhar2023perceiver,li2024manipllm,yang2023pave,jia2024lift3d,liu2024robomamba,xiong2024aic} can be ambiguous and brief, making it challenging for the robot to understand the tasks, or they can be overly detailed, increasing the difficulty for the model to follow. Goal images~\citep{zhong20233d,black2023zero,bousmalis2023robocat,lynch2020learning,jiang2022vima}, while providing an accurate target, often contain extraneous information irrelevant to the task, such as background elements and non-interactive objects.
Some methods use human demonstration videos~\citep{chane2023learning} or generated videos~\citep{black2023zero,du2023video,yang2023learning,du2024learning} to outline tasks step-by-step. However, human demonstration videos are burdensome to encode, and generated videos depend heavily on their quality. To address these challenges, several works propose using visual prompts as a convenient yet expressive modality for goal specification. These visual prompts are easy for users to create and can effectively convey the precise goals that the policy model should focus on.

\begin{figure*}[t]
\begin{center}
   \includegraphics[width=0.95\textwidth]{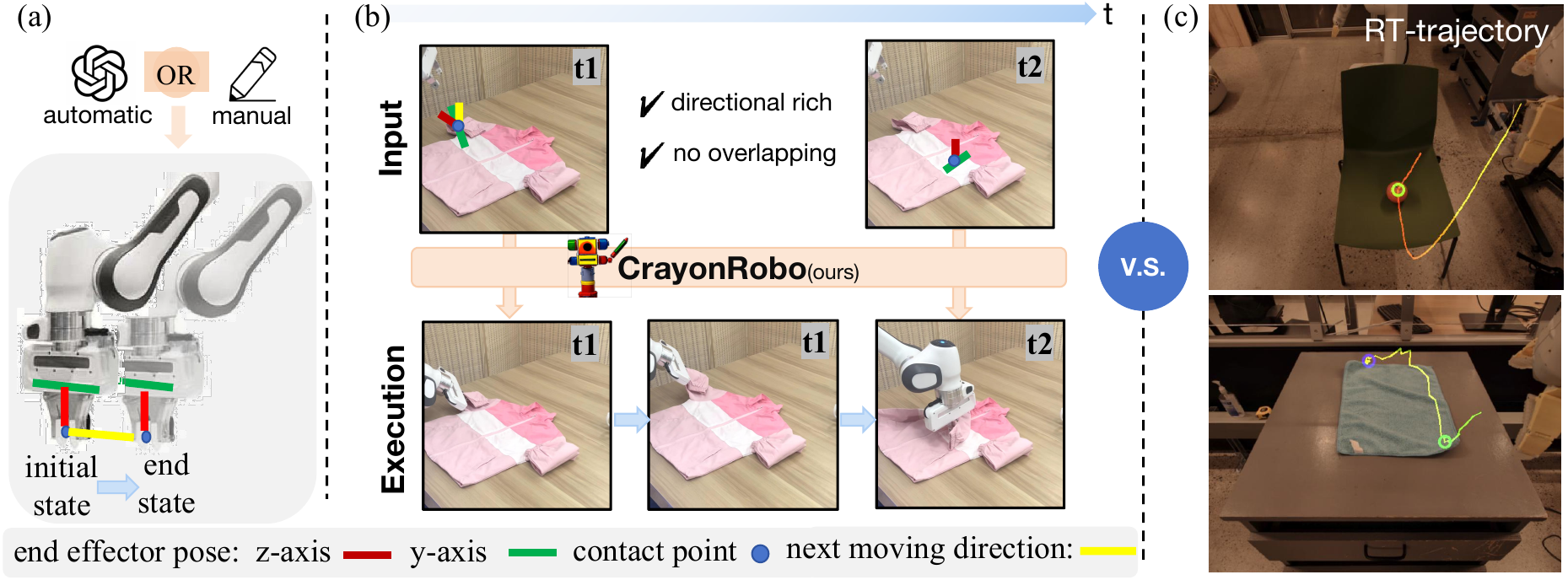}
   \caption{
   (a) shows our expression of different color prompts. (b) shows that we utilize a sequence of images with crayon visual prompts to express the key-frame steps, with each step illustrating the required low-level task goals, \emph{i.e.}, t1-pick, t2-place. Based on the input goal prompt, the model determines the 6 DoF contact pose, enabling it to interact with the object as required. When a yellow prompt is presented, the model also predicts 3D movement directions, guiding the movement after contact, \textit{e.g.}, picking upward in t1. For simple steps, such as t2-place, there is no need to present the post-contact moving direction. By sequentially executing each step in the key-frame sequence, the overall task is completed. (c) shows its differences with RT-trajectory~\citep{brohan2023rt}}
   \label{fig:intro}
\end{center}
\vspace{-0.8cm}
\end{figure*}

Among visual prompt-conditioned approaches, methods ~\citep{nasiriany2024pivot,gu2023rt,stone2023open,li2025hamster,yuan2024robopoint} have attempted to convey task goals more effectively. The RT-Sketch~\citep{sundaresan2023rt} method emphasizes drawing the target state of the most relevant object to represent the goal. However, it only depicts the final state and overlooks the intermediate key frames that are crucial for successful task execution. In contrast, RT-Trajectory~\citep{gu2023rt} (Figure\ref{fig:intro}.(c)) illustrates the entire movement path of the end-effector, which helps bridge the gap between task components and enhances generalization. Despite its advantages, they provide only positional information and neglect the action directional information, which is also critical for accurate task completion. 
Additionally, as trajectories become longer and overlap between key-frames, they can create confusion for the model regarding overall task planning.

This leads us to consider: \textit{is there any way to precisely and non-redundantly convey the goal while also clearly communicating the end-effector's action?}
% Therefore, we propose drawing or automatically generating crayon visual prompts on images to represent both low-level actions and high-level planning.
Since key-frames represent important or bottleneck steps of the gripper during the task execution~\citep{shridhar2023perceiver,goyal2023rvt,goyal2024rvt,johns2021coarse,yuan2023m2t2}, we propose \textit{\textbf{CrayonRobo}}, an approach that automatically generates or draws crayon-style visual prompts on images to effectively represent both low-level actions and high-level planning within key-frame sequences.
As illustrated in Figure \ref{fig:intro}(a), these prompts can include \textcolor{blue}{contact point}, \textcolor{red}{end-effector z-axis direction}, \textcolor{green}{end-effector y-axis direction}, and \textcolor{yellow}{next moving direction} after contact, conveying the objective of where and how to interact with the object.
% Observing constraints among directions, such as the relationship between the corresponding 2D directional prompts and 3D directions, 
We further develop a training strategy and supervision objectives aimed at progressively enhancing the model's ability to comprehend each type of prompt.
This allows the model to predict the corresponding SE(3) contact pose and the 3D next movement direction after contact, facilitating the completion of each sub-goal task.
Furthermore, as illustrated in Figure \ref{fig:intro}(b), we use key-frame sequences as input to express the task planning procedure, with each image presented with crayon visual prompts indicating its key-frame task goal.
We leverage the predicted 3D movement direction to connect each step sequentially, completing the overall long-horizon task.
This approach enables us to instruct the model's task objectives while also enhancing its robustness when encountering novel tasks by providing model-understandable prompts.

Our experimental setup includes a diverse range of manipulation tasks, both familiar and novel, where our method achieves a promising success rate in manipulation.
% In real-world scenarios, we validate the performance of our method on tasks that some may not be encountered during training.
Additional demonstrations are available at \href{https://sites.google.com/view/crayonrobo}{https://sites.google.com/view/crayonrobo}.

In summary, our contributions are as follows:
\begin{itemize}
\item 1) We propose employing a sequence of key-frames presented with prompts to explicitly convey the task objectives in both low-level action and high-level planning.
\item 2) We train a model that comprehends the visual-language prompts and predicts accurate contact poses along with moving directions in SE(3) space, ensuring the reliable completion of each key-frame task and the connection between steps to formulate a long-horizon task.
\item 3) Experiments at scale demonstrate its promising performance and robustness.
\end{itemize}
\section{Related Work}
\label{sec:related_work}
\textbf{Pure Vision-based Manipulation.}
In vision-based robotic manipulation, numerous studies have employed a variety of solutions, including deep learning~\citep{brohan2022rt,goyal2023rvt,shridhar2023perceiver,brohan2023rt}, imitation learning~\citep{chi2023diffusion,ze20243d,ke20243d,ju2024robo,liu2025hybridvla,huang2024manipvqa}, and reinforcement learning~\citep{an2024rgbmanip,luo2024serl,dai2023safe,nguyen2019review}.
For example, in deep learning-based solutions, some methods design action policy networks ~\citep{mo2021where2act,eisner2022flowbot3d,xu2022universal,wen2023any,bahl2023affordances} to calculate dense affordance maps and determine contact points and action poses.
However, pure vision-based manipulation networks focus more on actionability than functionality, potentially failing when functional actions are required to meet human needs. 
% To address this, more goal-oriented methods have been developed.

\noindent \textbf{Language-conditioned Manipulation.}
With advancements in language foundation models, language instructions are increasingly used for goal specification in goal-conditioned policy learning, as seen in~\citep{brohan2022rt}.
Other works, such as~\citep{liang2023code,ahn2022can,nair2022learning,lynch2020language,huang2023voxposer,shridhar2022cliport,shridhar2023perceiver,li2024manipllm,yang2023pave,xiao2022robotic}, utilize templated or freeform language for task specification.
Building on this, ~\citet{belkhale2024rt} introduces language motions as an intermediate layer between high-level goals and low-level actions.
Despite progress, challenges remain: language instructions often struggle to specify detailed actions or convey spatial objectives, frequently requiring human assistance to work effectively.

\noindent \textbf{Goal Image or Video-conditioned Manipulation.}
To enhance goal specificity and detail, several image-conditioned policy representations have been developed, with goal-image conditioning being one of the most prominent techniques. In goal-image conditioning, a final goal image defines the desired end state of a task~\citep{zhong20233d,black2023zero,bousmalis2023robocat,lynch2020learning,jiang2022vima}. This approach inputs both the initial and target states of the object and outputs the actions needed to achieve the goal.
In addition to static images, some methods use video or generated video~\citep{black2023zero,du2023video,yang2023learning,du2024learning} to represent each step of the process frame by frame, offering a more detailed execution procedure. 
However, both solutions encode excessive information, much of which may be irrelevant to the task, such as background details or unrelated objects.

\noindent \textbf{Visual Prompt-conditioned Manipulation.}
To address the issue of redundant goal information, recent approaches have proposed using visual prompts as goals. 
% These prompts provide concise, task-relevant information, which helps reduce unnecessary complexity and enhance task performance.
~\citet{sundaresan2023rt} suggests using goal sketches to indicate the target of the task-involved object, while ~\citet{gu2023rt} proposes drawing moving trajectories with key waypoints to specify the desired movement of the end-effector.
Additionally, ~\citet{nasiriany2024pivot} introduces a method that iteratively selects waypoints from a pool of potential options to form the overall trajectory, and ~\citet{yang2023pave,stone2023open,liu2024moka} use external models to choose waypoints based on mark-based visual prompts to complete tasks.
% While these approaches effectively focus on the robot's position or movement, they often overlook the importance of rotation, which is also crucial for task execution.
% Moreover, some methods consider the task as a whole, leading to potential visual prompt overlap as the task lengthens.

Different from them, we propose an innovative solution that allows manual drawing or automatic generating 2D prompts to represent both low-level action and high-level planning goals.
This approach clearly conveys the objective of each key-frame step, \emph{i.e.}, where and how to interact with the object, as well as the required task-planning procedure.
\section{Method}

% \subsection{Overview}
% In Section \ref{sec:task} and \ref{sec:data}, we describe task formulation and how the training data with crayon visual prompts was obtained. 
% Following that, in Section \ref{sec:train}, we explain how to train a model that can comprehend these 2D visual prompts and predict accurate 3D pose along with moving direction. 
% This step is crucial because the model's ability to accurately complete atomic tasks forms the foundation for performing long-horizon tasks.
% In Section \ref{sec:infer}, we elucidate how to obtain the required input during the inference stage and how to complete the entire manipulation tasks.

\begin{figure*}[ht]
% \vspace{-0.3cm}
\begin{center}
   \includegraphics[width=0.9\textwidth]{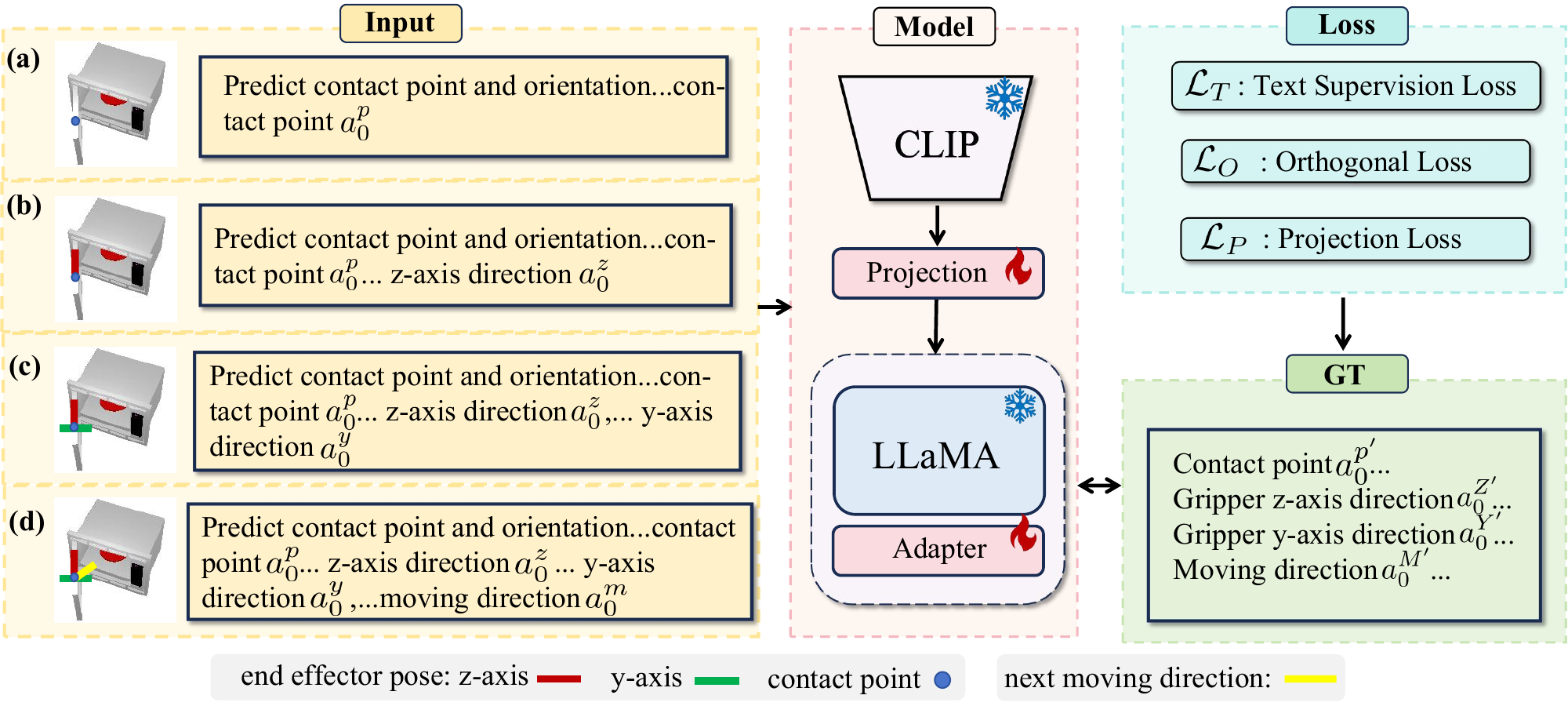}
   % \vspace{-0.2cm}
   \caption{We design training pairs that convey varying levels of information to enable the model to comprehend each type of prompt and introduce loss functions to guide it in predicting accurate poses.}
   \label{fig:policy}
   \vspace{-0.6cm}
\end{center}

\end{figure*}
% \subsection{Training Strategy}
% \label{sec:train}
 
% With the predicted contact pose and the moving direction after the approach, the robot is able to complete each atomic step.

\subsection{Problem Formulation}
\label{sec:task}
% The task is defined as follows: as shown in Figure 2, given a visual observation $I\in \mathbb{R}^{H\times W \times 3}$ of an object with crayon visual prompt, and a langugae prompt $P$ which describes the contact point coordinate\textcolor{blue}{$(x,y)$}, gripper 2D forward \textcolor{red}{$f$} and left \textcolor{green}{$l$} direction, and 2D moving direction \textcolor{yellow}{$m$}, the model generates the corresponding action $a_{0} = (a_{0}^{pos},a_{0}^{F},a_{0}^{L})$ and $a_{0}^{M}$, where $a_{0}^{pos}\in \mathbb{R}^{2}$  which is then projected to 3D action by depth map while all $a_{0}^{F}$, $a_{0}^{L}$, and $a_{0}^{M}$ belong to $\mathbb{R}^{2}$,  a unit vector in 3D indicating the end-effector forward, left, and moving direction.

The task is defined as follows: 
the model takes the visual input $I\in \mathbb{R}^{H\times W \times 3}$ of an object presented with crayon visual prompts. 
This prompt utilizes four colors: blue, representing the contact point; red and green, indicating the z-axis and y-axis directions of the end-effector when contacting the object; and yellow, symbolizing the next moving direction after contacting. 
To eliminate ambiguities caused by visual prompts overlapping each other, we specify their numerical values in text inputs as a language prompt $P$, including the coordinates of the contact point denoted as \textcolor{blue}{$a_{0}^{p}\in \mathbb{R}^{2}$}, the gripper's 2D z-axis direction \textcolor{red}{$a_{0}^{z}$} and y-axis direction \textcolor{green}{$a_{0}^{y}$}, and the 2D moving direction \textcolor{yellow}{$a_{0}^{m}$}, all represented as unit vectors in $\mathbb{R}^{2}$. 
%Following the settings of previous manipulation works that utilize depth maps for contact point projection~\citep{liu2024robomamba,li2024manipllm,mo2021where2act, xiong2024aic,huang2024manipvqa,huang2024a3vlm}, 
The objective of the model is to generate an action $a_{0} = (a_{0}^{p'},a_{0}^{Z},a_{0}^{Y},a_{0}^{M})$, where $a_{0}^{p'}$ is mapped into 3D position $a_0^P$ using the depth information and camera parameters, while \textcolor{red}{$a_{0}^{Z}$}, \textcolor{green}{$a_{0}^{Y}$}, and \textcolor{yellow}{$a_{0}^{M}$} are unit vectors in $\mathbb{R}^{3}$ representing the end-effector's 3D z-axis, y-axis, 
and moving directions, respectively. 
$a_{0}^{Z}$ and $a_{0}^{Y}$ jointly determine the rotation for contacting the object, while $a_{0}^{M}$ controls the following movement after contact. 
Note that, the prediected $a_{0}^{p'}$ can be different from the $a_{0}^{p}$ in the text prompt input due to potential input noise. 
% Meanwhile, given the inherent ambiguity of low-dimensional representations, the same 2D prompt input can correspond to multiple high-dimensional output probabilities. The model must therefore predict the correct output by interpreting detailed local object information across diverse object categories and adapting to the varying perspectives of different camera viewpoints.

\subsection{Data Collection}
\label{sec:data}
In this section, we elaborate on how to obtain the 2D prompt required for training. In the simulator, building upon some effective rule-based heuristic approaches~\citep{mo2021where2act,li2024manipllm}, we interact with objects and record the success contact pose of the end-effector including 3D contact point, z-axis direction $a_{0}^{Z'}$, and y-axis direction $a_{0}^{Y'}$ along with the moving direction $a_{0}^{M'}$ of the object contact point.
These are then served as ground-truth to guide the training.

As for the input prompt, 3D contact point is unprojected onto a 2D image as $a_0^p =(x,y)$.
We then adopt 3D contact point as the center and locate other 3D points along $a_{0}^{Z'}$ and $a_{0}^{Y'}$. These points are then unprojected onto the 2D image. Subsequently, we connect these points with $a_0^p$ to calculate 2D vectors $a_{0}^{z}$ and $a_{0}^{y}$, while obtaining red and green lines respectively.
We obtain the yellow moving direction line and obtain the 2D moving direction vector $a_{0}^{m}$ in the same way utilizing $a_{0}^{M'}$.
These are then served as information embedded in the input language prompt.
The corresponding visual prompts are generated by overlaying these elements in their respective colors onto the images.
% To be robust to the potential noise on the prompt input in real world, we add noise on both positional and directional prompts during training.
% Specifically, $a_{0}^{p}$ in the input prompt denotes adding noise on $a_{0}^{P'}$ for more robust model training.

\subsection{Training Strategy}
\label{sec:train}
% Given both the visual and language prompts, we require the model to effectively interpret inputs from both modalities. Given the robust language understanding and visual processing capabilities of Vision Language Action Models (VLAs) and inspired by their applications in prior robotic manipulation tasks ~\citep{li2024manipllm,li2023vision,liu2024robomamba,kim2024openvla}, we have selected VLAs as the backbone of our approach.
% In this section, we demonstrate how we enable models to comprehend 2D prompts and equip them with manipulation capabilities.
% As shown in Figure \ref{fig:policy}, we design fine-tuning tasks for VLAs by creating pairs of inputs with varying levels of information and crafting loss functions to guide the policy training. By doing so, the model can effectively execute the required low-level actions, \emph{i.e.}, where to contact and how to interact. Thus, when faced with novel objects, as long as these model-interpretable prompts are provided, the model can predict accurate actions, enhancing its robustness.

\subsubsection{Model Architecture}
\label{sec:model}
Given both the visual and language prompts, we require the model to effectively interpret inputs from both modalities. Observing the robust language understanding and visual processing capabilities of Vision Language Action Models (VLAs) and inspired by their applications in prior robotic manipulation tasks ~\citep{li2024manipllm,li2023vision,liu2024robomamba,kim2024openvla}, we follow the backbone architecture of an open-source VLA model~\citep{li2024manipllm} and adopt its parameter updating strategy which only finetunes the injected adapters~\citep{hu2021lora} within LLaMa~\citep{touvron2023llama}, alongside the multi-modal projection module, while keeping the primary parameters frozen. This approach aims to preserve the inherent existing pre-trained knowledge, particularly in sim-to-real transfer, while enhancing the model's ability to comprehend visual prompts and perform robotic manipulation in an efficient training time.
% Note that, the inference speed of this backbone is 5-6 HZ under 128 output tokens.
Specifically, when presented with an RGB image containing visual prompts $I$, we employ CLIP's visual encoder~\citep{radford2021learning} to extract visual features. Simultaneously, text prompts $P$ are encoded into text features using LLaMa's pre-trained tokenizer~\citep{touvron2023llama}. The alignment of visual and text feature representations is achieved through the multi-modal projection module.

\subsubsection{Policy Learning}
\label{sec:policy}

\textbf{Input pairs: } To help the model better understand the meaning of various prompts, we design a series of visual-textual input pairs that progressively convey different levels of information. This gradual progression enables the model to develop a deeper understanding of the physical significance of each component. 
Specifically, as illustrated in (a)-(d) in Figure \ref{fig:policy} input. We formulate the problem of pose prediction as a language modeling task, directly outputting 3D directions in textual form. Thus, we generate the ground-truth text as shown in Figure \ref{fig:policy} to guide the training.
The entire language description is shown in Appendix.~\ref{sec:A1}.

% \subsubsection{Objective Functions}
\noindent \textbf{Training objectives: } Our model's objective is to accurately predict 3D pose based on 2D visual-language prompts given different object and camera views. Therefore, we introduce the following losses to guide the policy training:

\textit{Text Supervision Loss $\mathcal{L}_{T}$:}
This loss ensures the effective alignment of the model’s visual and linguistic input, making sure it can output the correct output pattern.
We formulate pose prediction as a classification task by discretizing the continuous numbers in the normalized 3D direction vector into 100 discrete bins [-50,50], with each bin spanning 0.02, and adopt cross-entropy loss to supervise. 

\textit{Orthogonal Loss $\mathcal{L}_{O}$:}
The orthogonality of a rotation matrix necessitates that its components maintain orthogonal relationships between each pair of directions. However, formulating direction prediction as language predictions for $a_{0}^{Z}$ and  $a_{0}^{Y}$ does not explicitly consider their geometric relation. To address this, we extract $a_{0}^{Z}$ and $a_{0}^{Y}$ from the model's output text and ensure their orthogonal relationship under Gram-Schmidt loss~\citep{giraud2005loss}.
% : $\mathcal{L}_{O} =\left|\frac{\mathbf{a_{0}^{Z}}\cdot\mathbf{a_{0}^{Y}}}{\|\mathbf{a_{0}^{Z}}\| \|\mathbf{a_{0}^{Y}}\|}\right|$
% \[
% \mathcal{L}_{O} =\left|\frac{\mathbf{a_{0}^{Z}}\cdot\mathbf{a_{0}^{Y}}}{\|\mathbf{a_{0}^{Z}}\| \|\mathbf{a_{0}^{Y}}\|}\right|
% \]

\textit{Projection Loss  $\mathcal{L}_{P}$:}
3D directional predictions should align with their corresponding 2D directional prompts when unprojected back to the 2D plane.
To establish an explicit connection between the input 2D directional prompt and the output 3D directions, we introduce a projection loss designed to guide their correlation.
Specifically, in the model's output, we extract the contact point $a_{0}^{P}$ and map it to 3D space using the depth map and camera parameters. Subsequently, leveraging the predicted 3D direction in the z-axis $a_{0}^{Z}$, we locate 3D points along the direction with the contact point as the center. 
The 3D point are then unprojected back onto the 2D plane, and connecting with $a_{0}^{P}$ to generate predicted 2D directional vectors $a_{0}^{z*}$.
We employ a similar approach to acquire $a_{0}^{y*}$ and $a_{0}^{m*}$, which are then supervised based on the 2D directional prompts $a_{0}^{z}$, $a_{0}^{y}$, and $a_{0}^{m}$ using cosine similarity loss
, where \( \cdot \) denotes the dot product: 
% $\mathcal{L}_{P} = (1 - \frac{\mathbf{a_{0}^{z'}}\cdot\mathbf{a_{0}^{z}}}{\|\mathbf{a_{0}^{z'}}\| \|\mathbf{a_{0}^{z}}\|}) + (1 - \frac{\mathbf{a_{0}^{y'}}\cdot\mathbf{a_{0}^{y}}}{\|\mathbf{a_{0}^{y'}}\| \|\mathbf{a_{0}^{y}}\|})
%  + (1 - \frac{\mathbf{a_{0}^{m'}}\cdot\mathbf{a_{0}^{m}}}{\|\mathbf{a_{0}^{m'}}\| \|\mathbf{a_{0}^{m}}\|})$
{\small
\[
\mathcal{L}_{P} = (1 - \frac{\mathbf{a_{0}^{z*}}\cdot\mathbf{a_{0}^{z}}}{\|\mathbf{a_{0}^{z*}}\| \|\mathbf{a_{0}^{z}}\|}) + (1 - \frac{\mathbf{a_{0}^{y*}}\cdot\mathbf{a_{0}^{y}}}{\|\mathbf{a_{0}^{y*}}\| \|\mathbf{a_{0}^{y}}\|})
 + (1 - \frac{\mathbf{a_{0}^{m*}}\cdot\mathbf{a_{0}^{m}}}{\|\mathbf{a_{0}^{m*}}\| \|\mathbf{a_{0}^{m}}\|})
\]
}

Note that, $\mathcal{L}_{T}$ and $\mathcal{L}_{O}$ are consistent across all input pairs, but $\mathcal{L}_{P}$ varies based on the input prompts. 
For example, in Input(c) of Figure.~\ref{fig:policy} where does not include the $a_{0}^{m}$, we will exclude the supervision for $a_{0}^{m}$ and $a_{0}^{m*}$ in $\mathcal{L}_{P}$.
After completing the training process, the model, having been trained with dynamic input patterns, can understand each type of prompt and predict the required actions based on the provided prompts.
The aforementioned losses are trained simultaneously under the total objective function: $\mathcal{L}$ = $\lambda_{1}*\mathcal{L}_{T}$+$\lambda_{2}*\mathcal{L}_{O}$+$\lambda_{3}*\mathcal{L}_{P}$.

% \subsection{Inference Stage}

% \textit{Human drawings}: 
% Utilizing human drawings is an intuitive and practical approach to creating trajectory sketches. To efficiently generate these sketches, we have developed a straightforward graphical user interface (GUI) that allows users to draw crayon visual prompts based on the initial camera image captured by the robot. The visual prompts created can include the following elements: 1）contact point (blue dot), 2) contact point + forward direction vector (blue dot and red line), 3) Contact point + forward direction vector + left direction component (blue dot, red line, and green line). This provides users with more input options.

% \textit{LLM prompting}:
% In order to automate the entire process, we also attempted to use off-the-shelf models to generate these crayon visual prompts. However, we found that it's challenging to find off-the-shelf models that can provide hints for 2D directional vectors. Nonetheless, some off-the-shelf models excel in providing suitable 2D contact points based on text prompts. We utilized ManipVQA, which considers task descriptions and object affordances, to output appropriate 2D contact points. This output can serve as input prompts for our method regarding the contact point.
\subsection{Inference and Interaction}
\label{sec:infer}
% \label{sec:infer}
\begin{figure*}[h]
\begin{center}
   \includegraphics[width=0.90\textwidth]{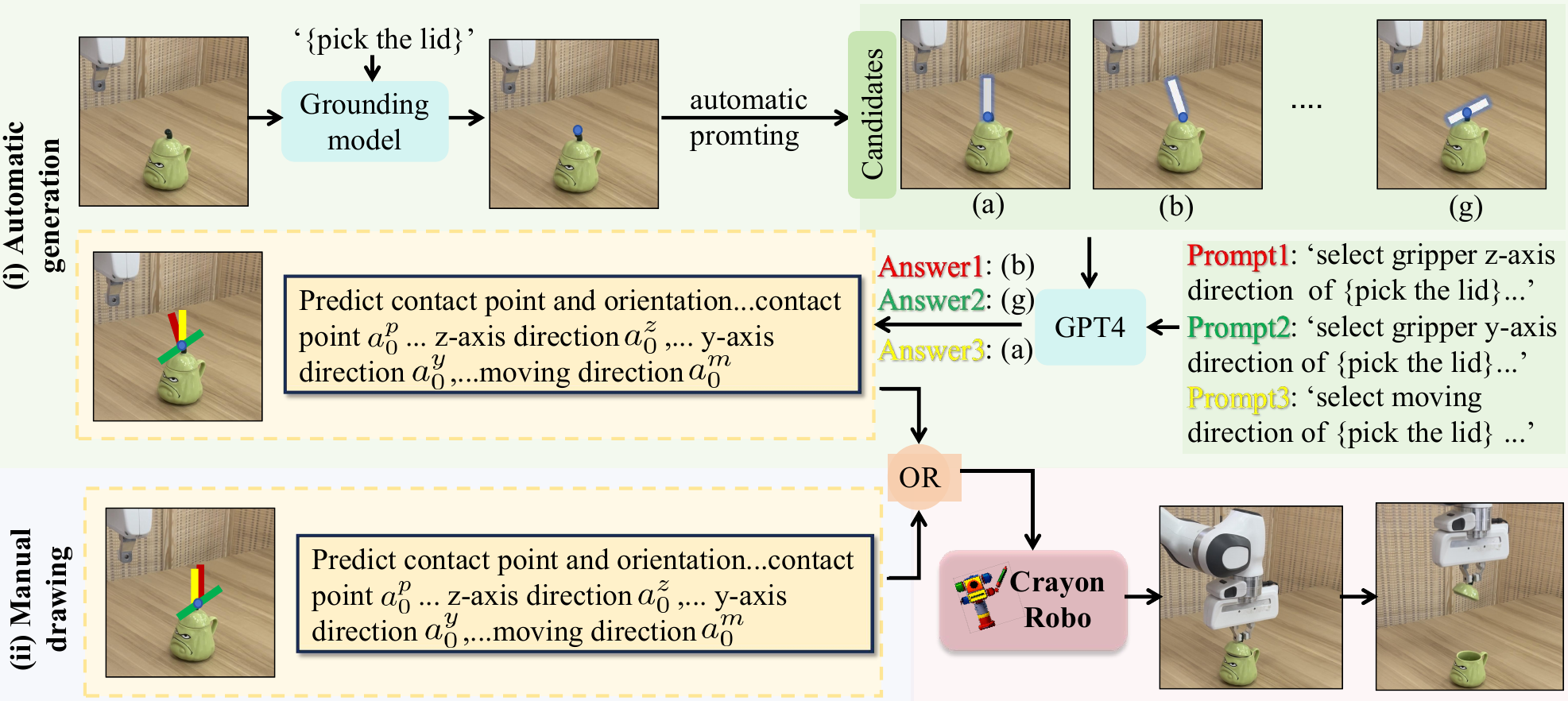}
   \caption{
Illustration of model inference with input generated in different ways. 
}
   \label{fig:infer}
\end{center}

\end{figure*}
\subsubsection{Model Inference}
During the inference stage, we provide both automated and manual drawing approach to extract these visual prompts, as illustrated in Figure \ref{fig:infer}(i) and (ii). 
For automatic generation approach, we first use Grounded-DINO \citep{liu2023grounding} to detect the object's bounding box and select its center, forming the blue circle. Then, we automatically generate 32 uniformly sampled 2D directional lines around the full 360-degree circle with the blue circle as the center or as the middle. GPT-4 \cite{achiam2023gpt} is then prompted to select lines from all candidates to represent the gripper's z-axis direction, y-axis direction, and moving direction, resulting in the red, green, and yellow lines, respectively.
As shown in Figure \ref{fig:infer}(ii), for manual drawing approach, we allow users to draw crayon visual prompts on the image, which then serve as the visual input. 
% The blue, red, and green prompts indicate the pose that the end effector should reach, while the yellow line represents the moving direction after contact. Meanwhile, we also

% The key-frames represent important or bottleneck steps of the gripper during the task execution [55], such as a grasp, or place pose. 
The contact pose is always required, while whether the movement direction after contact is needed depending on the task characteristics.
Specifically, \citet{shridhar2023perceiver,goyal2023rvt,goyal2024rvt,johns2021coarse,yuan2023m2t2} observe that long-horizon tasks usually can be decomposed into several key-frame tasks. 
Meanwhile, different key-frame tasks exhibit distinct motion patterns: some require a specific moving direction to guide the subsequent movement after contact, while others end when executing contact pose.
For example, \textit{pick} action requires moving along $a_{0}^{M}$, which usually indicates an upward movement. Similarly, \textit{pushing} and \textit{pulling} actions require the robot to determine the direction in which to push or pull.
In contrast, certain primitives, such as \textit{place} and \textit{move}, are simple motions that do not require additional movement after approaching the object. For example, given a pick-and-place task, once the object is picked up, the ``pre-place'' pose is already established. Once the object is placed according to the contact pose prompt, the task is considered complete. Therefore, there is no need to draw a moving prompt for these primitives.
For actions like \textit{rotate}, such as rotating a button, the movement typically does not involve translating the position after contact. Instead, it focuses on rotating the last joint. As a result, $a_{0}^{m}$, which specifies translation, is not needed. Instead, we provide two key-frames as prompts: one depicting the end effector pose at the moment of contact and the other after the rotation. From these two temporal prompts, the required rotation direction becomes evident.
More visual details are demonstrated in the Appendix.~\ref{sec:A4}.
After generating the visual prompt image, we automatically extract the coordinates of the contact point and the direction vectors based on the pixel RGB values and then incorporate them into the language prompt. Given the visual and language input, the model outputs the predicted action $a_{0}$.

% In contrast, some primitives require the moving direction to determine the moving action. For example, \textit{pick} action requires moving along $a_{0}^{M}$, which is usually an upward movement. For both \textit{pushing} and \textit{pulling}, the robot must determine the direction in which to push or pull. Thus, for different primitive actions, the user should utilize different prompt patterns to specify the action goal, with or without moving direction prompts.
Note that although 2D directions may overlap, the model does not require them to originate from the same point, as long as the direction is accurate and clearly visible in the image, the model can interpret it. Additionally, if the direction vector is perpendicular to the camera, it may be challenging to represent it as a 2D projection. In such cases, adding slight noise to the direction vector can enhance its visibility. Since the model is trained to interpret such situations, it can handle similar scenarios during inference, where the direction may be perpendicular to the camera.
% Regardless of whether it starts from the contact point, the extracted directional text prompt remains the same.

% After generating the visual prompt image, we automatically extract the coordinates of the contact point and the direction vectors based on the pixel RGB values and then incorporate them into the language prompt. Given the visual and language input, the model outputs the predicted action $a_{0} = (a_{0}^{P},a_{0}^{Z},a_{0}^{Y},a_{0}^{M})$.

\subsubsection{Interaction Strategy}
We illustrate how we enable the robot to interact with objects given the predicted action $a_{0}$.
Same with other VLA models~\citep{liu2024robomamba,huang2024a3vlm,li2024manipllm}, $a_{0}^{p}$ is projected into 3D space utilizing depth information and camera parameters. The $a_{0}^{Z}$ and $a_{0}^{Y}$ jointly contribute to determining the rotation matrix of the end-effector, facilitating the establishment of initial contact with the object. If provided with $a_{0}^{m}$, we then follow the predicted moving direction $a_{0}^{M}$ to determine the subsequent movements after contact. 

For the long-horizon tasks, we leverage a sequence of key-frames with visual prompts to serve as high-level planning, with each frame representing a sub-goal step. By executing the key-frame steps sequentially, we can complete the long-horizon tasks. The benefit of this approach revolves around breaking down the complexity of long-horizon tasks and allowing us to optimize the success rate of each key-frame step to ensure the overall task's success. Meanwhile, regarding each key-frame step as compositions, we can group them into arbitrary combinations, enabling the model to handle diverse manipulation tasks.
\section{Experiment}
\label{sec: exp}
In our experiments, we mainly focus on exploring the
following questions:
\begin{itemize}
    % \item Section \ref{sec:baseline}: Is the method robust to novel tasks and different camera view angles? 
    \item Section \ref{sec: prompt type}: What is the effect of different types of prompts on model performance?
    \item Section \ref{sec: noise}: How does the model handle noise in the input prompt, and is it user-friendly?
    \item Section \ref{sec:real}: Can the model perform specific tasks without prompts, given certain input distributions?
    % \item What is the performance when there are noises on the visual prompt?
\end{itemize}
\subsection{Setup Details}
% \usepackage[table]{xcolor}
% \textbf{Data Collection.}
% We adopt SAPIEN~\cite{Xiang_2020_SAPIEN} and the PartNet-Mobility dataset to set up an interactive environment for our task, with VulkanRenderer of high-efficiency rasterization-based renderer.
% We use a Franka Panda Robot with finger gripper as the robot actuator.
% We sample the training data offline with approximately 10,000 manipulation success samples across 25 categories. 
% We randomly select a contact point $p$ on the movable part and the rotation matrix for interacting with the object. If successful manipulation is achieved, we record it as a successful sample. 
% % The category information required for training can be obtained from the simulator, and affordance prior is collected following the method described in Section.\ref{sec:ft}. 
% For tasks involving pulling action, it requires the movement direction of the object part and the opposite of the end effector direction to be within the same hemisphere. For tasks involving pushing action, 

% \textbf{Training Details.}
% We fine-tune the LLaMA-Adapter~\cite{zhang2023llama} using an 80G A100 GPU, leveraging approximately 10,000 manipulation success samples over 10 epochs with a batch size of 4, requiring a training time of 2 hours. 
% The model incorporates a pre-trained CLIP-ViT-L~\cite{radford2021learning} model, a 7B LLaMA~\cite{touvron2023llama} model, and a multi-modal projection module comprising 32 transformer layers.

\textbf{Data Collection.}
Following previous work~\citep{mo2021where2act,li2024manipllm}, we utilize SAPIEN~\cite{Xiang_2020_SAPIEN} along with the PartNet-Mobility dataset to construct an environment, interacting with about 1500 object shapes under various camera view randomization. The tasks involve several actions that require changes in both translation and rotation,  \textit{e.g.}, pull the drawer, open the door, open the laptop lid, etc. The setup uses the flying Franka Panda Gripper for both the training and testing stages. We follow the procedure in Section \ref{sec:data} to collect prompts within the simulator, which takes about 6-8 hours to collect about 10,000 training samples. 
Details of the training and testing dataset split and camera view randomization can be found in Appendix.~\ref{sec:A2}.

\noindent \textbf{Evaluation Metric.}
We utilize the manipulation success rate to assess the effectiveness of the manipulation, calculated as the ratio of successfully manipulated samples to the total number of test samples. A successful sample is defined using a binary criterion, with success determined by thresholding the distance that the object part moves.
% To evaluate the accuracy of predicted 3D directions, we employ cosine similarity, comparing predicted directions with ground-truth directions.
% Both initial and long-distance movements apply an active impedance adaptation policy to adjust movement direction.

\begin{table*}[h]
% \vspace{-0.4cm}   
 
	\begin{center}
	
	\small
	    \setlength{\tabcolsep}{0.8mm}{
		\begin{tabular}{c| cc c c c c c c c c c c c c c c}
  \hline
	\multirow{2}{*}{\textbf{}}&\multirow{2}{*}{\textbf{}} &\multicolumn{15}{c}{\textbf {Seen Tasks}}\\
 Method
    & \includegraphics[width=0.033\linewidth]{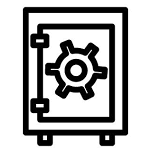}
    &\includegraphics[width=0.033\linewidth]{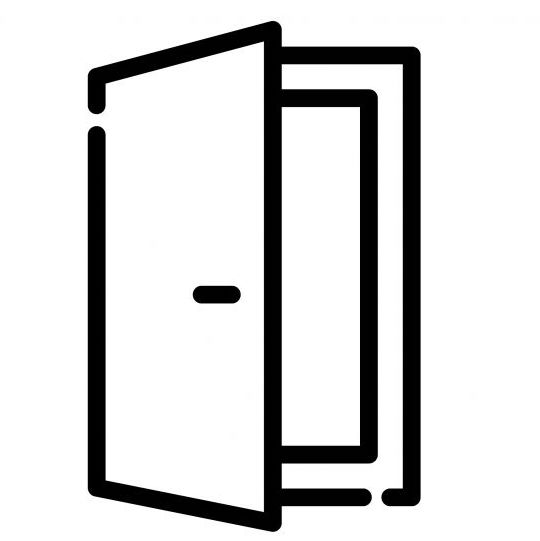}
        &\includegraphics[width=0.033\linewidth]{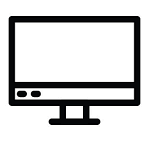}
        &\includegraphics[width=0.033\linewidth]{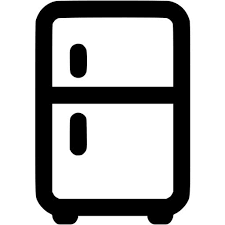}
        &\includegraphics[width=0.033\linewidth]
        {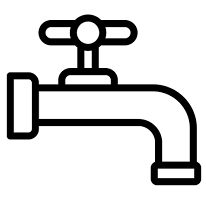}
        &\includegraphics[width=0.033\linewidth]
        {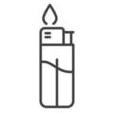}
        &\includegraphics[width=0.033\linewidth]{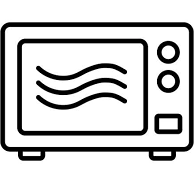}
        &\includegraphics[width=0.033\linewidth]{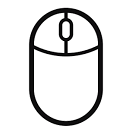}
        &\includegraphics[width=0.033\linewidth]{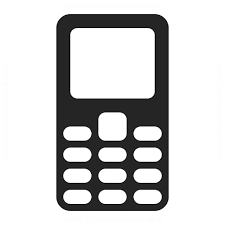}
        &\includegraphics[width=0.033\linewidth]{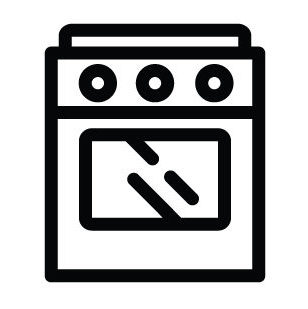}
        &\includegraphics[width=0.033\linewidth]{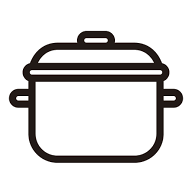}
        &\includegraphics[width=0.033\linewidth]{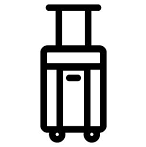}
        &\includegraphics[width=0.033\linewidth]{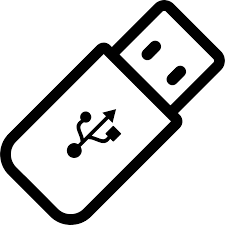}
        &\includegraphics[width=0.033\linewidth]{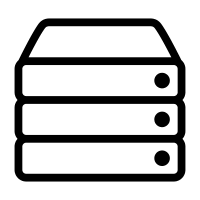}
        &\includegraphics[width=0.033\linewidth]{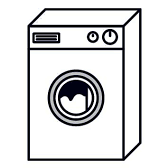}
        &\includegraphics[width=0.033\linewidth]{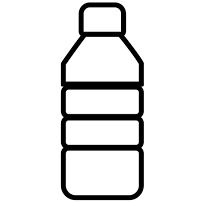}\\\hline\hline
        
        Flowbot &0.76 &\textbf{0.86}& 0.08& 0.67& 0.26& 0.05& 0.58& 0.29& 0.71 &0.35& 0.07& 0.23& 0.40& 0.63& 0.52 &0.04 \\
        
        ManipLLM&0.68&0.64	&0.36&	0.77&	0.26&	0.62&	0.65&	0.61&	0.38&	0.52&	0.53&	0.40&	0.64&	0.71&	0.83&	0.64  \\
  Implicit3D& 0.53&0.58&0.35&0.55&0.30&\textbf{0.66}&0.58&0.51&0.31&0.41&0.45&0.34&0.69&0.54&0.31&0.43\\  
 \rowcolor{verylightgrey} Ours(s)& \textbf{0.89}&0.85&\textbf{0.67}&\textbf{0.91}&\textbf{0.87}&0.50&\textbf{0.92}&\textbf{0.85}&\textbf{0.87}&\textbf{0.85}&\textbf{0.54}&\textbf{0.84}&\textbf{0.85}&\textbf{0.92}&\textbf{0.85}&\textbf{0.75}\\\hline
  RT-Traj&0.56&0.58&0.28&0.45&0.56&\textbf{0.50}&0.65&0.42&0.29&0.56& 0.40&0.49&0.46&0.53&0.52&0.27 \\

%  \rowcolor{gray!20}
  % \rowcolor{verylightgrey}Ours(a)& \textbf{0.71} & \textbf{0.79} & \textbf{0.58} & \textbf{0.81} & \textbf{0.79}&0.43&\textbf{0.81}&\textbf{0.81}&\textbf{0.75} & \textbf{0.78}&\textbf{0.50}&\textbf{0.75}&\textbf{0.76}&\textbf{0.79}&\textbf{0.81} &\textbf{0.73} \\
  \rowcolor{verylightgrey}Ours(f)& \textbf{0.71} & \textbf{0.79} & \textbf{0.58} & \textbf{0.81} & \textbf{0.79}&0.43&\textbf{0.81}&\textbf{0.81}&\textbf{0.75} & \textbf{0.78}&\textbf{0.50}&\textbf{0.75}&\textbf{0.76}&\textbf{0.79}&\textbf{0.81} &\textbf{0.73} \\\hline
%   Ours($f_{l}$)& 0.71&0.71&0.45&0.79&0.71&0.43&0.80&0.66&0.71&0.78&0.44&0.73&0.64&0.72&0.70&0.59\\
% \hline 
  %\rowcolor{gray!20} 
 
	\multirow{2}{*}{\textbf{}}&\multirow{2}{*}{\textbf{}}&\multicolumn{8}{c|}{\textbf {Seen Tasks} }&

			\multicolumn{6}{c}{\textbf {Unseen Tasks}}\\
			
	Method
    & \includegraphics[width=0.033\linewidth]{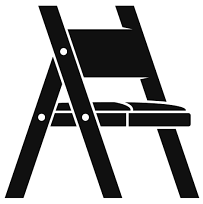}
        &\includegraphics[width=0.033\linewidth]{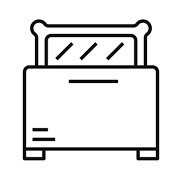}
        &\includegraphics[width=0.033\linewidth]{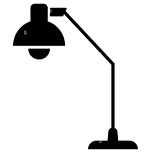}
        &\includegraphics[width=0.033\linewidth]{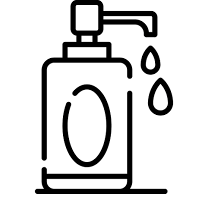}
        &\includegraphics[width=0.033\linewidth]{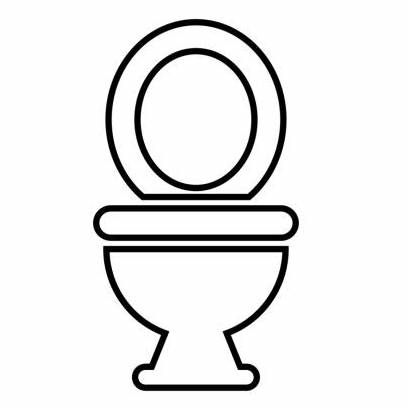}
        &\includegraphics[width=0.033\linewidth]{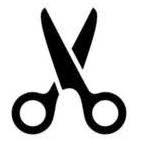}
        &\includegraphics[width=0.033\linewidth]{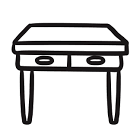}
        &\includegraphics[width=0.033\linewidth]{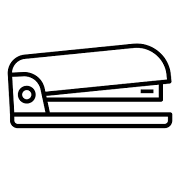}
         &\multicolumn{1}{c|}{{\textbf {AVG}} }
        &\includegraphics[width=0.033\linewidth]{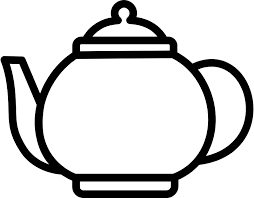}
        &\includegraphics[width=0.033\linewidth]{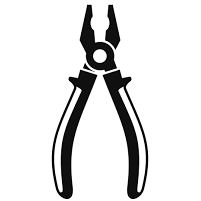}
        &\includegraphics[width=0.033\linewidth]{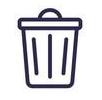}
        &\includegraphics[width=0.033\linewidth]{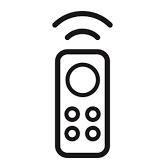}
        &\includegraphics[width=0.033\linewidth]{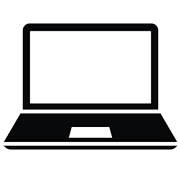}
        &\includegraphics[width=0.033\linewidth]{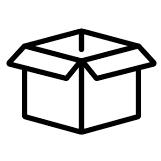}
        &{\textbf {AVG}}\\\hline\hline
 
  % \multicolumn{1}{|c|}{} & \multicolumn{1}{c|}{|} & \multicolumn{1}{c|}{|} & \multicolumn{1}{c|}{|} \\ % Add vertical lines in the last few rows 
Flowbot & 0.18& 0.10& 0.39& 0.10& 0.10& 0.61& 0.34& 0.19 &\multicolumn{1}{c|}{0.43}& 0.12& 0.60& 0.21& 0.51& 0.10& 0.22& 0.38\\

  ManipLLM&0.41&0.75&	0.44&	0.67& 0.38	&0.22&	0.81&\textbf{0.86}&	\multicolumn{1}{c|}{0.51}&	0.38&	0.85	&0.42	&0.60&	0.43&	\textbf{0.65}&	0.47\\
   Implicit3D&0.27&0.65&0.20&0.33&0.45&0.17&0.80&0.53&\multicolumn{1}{c|}{0.55} &0.15&0.41&0.57&0.39&0.28&0.52& 0.39\\  
  \rowcolor{verylightgrey} Ours(s)& \textbf{0.65}&\textbf{0.90}&\textbf{0.78}&\textbf{0.79}&\textbf{0.68}&\textbf{0.91}&\textbf{0.94}&0.46& \multicolumn{1}{c|}{\textbf{0.80}}&\textbf{0.47}&\textbf{0.88}&\textbf{0.83}&\textbf{0.70}&\textbf{0.76}&0.63&\textbf{0.79} \\ \hline
  RT-Traj&0.32&0.46&0.40&0.31&0.37&0.68&0.58&\textbf{0.47}&\multicolumn{1}{c|}{0.57} & 0.25&0.39&0.48&0.57&0.13&0.59&0.52 \\
  % \rowcolor{verylightgrey} Ours(a)& \textbf{0.63}&\textbf{0.67}&\textbf{0.48}&\textbf{0.78}&\textbf{0.67}&\textbf{0.74}&\textbf{0.83}&0.44&\multicolumn{1}{c|}{\textbf{0.74}}& \textbf{0.46}&\textbf{0.88}&\textbf{0.82}&\textbf{0.63}&\textbf{0.61} & \textbf{0.61} & \textbf{0.72} \\
  %\rowcolor{gray!20}
  
  \rowcolor{verylightgrey} Ours(f)& \textbf{0.63}&\textbf{0.67}&\textbf{0.48}&\textbf{0.78}&\textbf{0.67}&\textbf{0.74}&\textbf{0.83}&0.44&\multicolumn{1}{c|}{\textbf{0.74}}& \textbf{0.46}&\textbf{0.88}&\textbf{0.82}&\textbf{0.63}&\textbf{0.61} & \textbf{0.61} & \textbf{0.72} \\\hline
  % Ours($f_{l}$)& 0.52&0.63&0.39&0.70&0.55&0.73&0.77&0.41&\multicolumn{1}{c|}{0.69}&0.38&0.84&0.77&0.63&0.60&0.52&0.68\\ \hline 
  %\rowcolor{gray!20} 

		\end{tabular}
		% \vspace{-0.1cm}
	
 % \vspace{-0.1cm}}
 }
 \end{center}
% \vspace{-0.3cm}
\caption{Comparison of our method against baseline methods. (s) and (f) denote suction gripper and finger gripper, respectively. Bold text indicates the highest score within each end-effector type.}
\label{tab:main}

% 	\vspace{-0.25cm}
\end{table*}
\subsection{Comparisons with Baselines}

\label{sec:baseline}
% We compare CrayonRobo against three baseline methods: UMPNet~\cite{xu2022universal}, Flowbot3D~\cite{eisner2022flowbot3d}, and ManipLLM~\cite{li2024manipllm}. 
To ensure fairness in comparison, all methods adhere to the same train and test split.
% Notably, all these approaches employ a suction gripper as the end-effector. Therefore, to facilitate a direct and transparent comparison, we present the results of our method using both suction gripper Ours(s) and finger gripper Ours(f) in Table \ref{tab:main}. 

\textit{Pure vision model} Flowbot3D~\citep{eisner2022flowbot3d}:
It predicts motion direction on the point cloud, denoting it as \textit{flow}. The point with the largest flow magnitude serves as the interaction point, while the direction of the flow represents the end-effector's orientation and moving direction.

\textit{Language conditioned model} ManipLLM~\citep{li2024manipllm}:
It takes the task description and the object's initial image, and then outputs the end-effector pose.
We use its predicted contact point and rotation to contact with the object and adopt its active adaptation policy for moving. 

\textit{Vision Goal conditioned 3D model} Implicit3D~\citep{zhong20233d}:
It develops a manipulation policy that utilizes the transporter to detect key-points for 3D objects. By providing the initial and target state point cloud of the object, key points are then used to determine the end-effector pose for manipulation.

The above three methods use suction gripper as the end-effector, which is compared with ours(s) in Table. \ref{tab:main}

\textit{Visual prompt conditioned model} RT-trajectory~\citep{gu2023rt}: Since the RT-Trajectory code is not publicly available at the time of this paper's submission, we replicate its method for comparisons with implementation details in Appendix.~\ref{sec:A3}. 

\textit{Goal video conditioned model} AVDC~\citep{yang2023learning}: The generative model is trained on our training split. During testing, we employ Gemini-1.5~\citep{team2023gemini} to evaluate the quality of the generated videos by assessing whether they successfully complete the given tasks, \textit{e.g., }open the door. The ratio of successfully generated tasks is 10.2\%. Given that the performance of the generative video-based execution policy is highly dependent on the quality of the generated videos, we believe this score can somehow reflect its overall effectiveness during the following execution.

\begin{figure*}[h]
\begin{center}
   \includegraphics[width=0.9\textwidth]{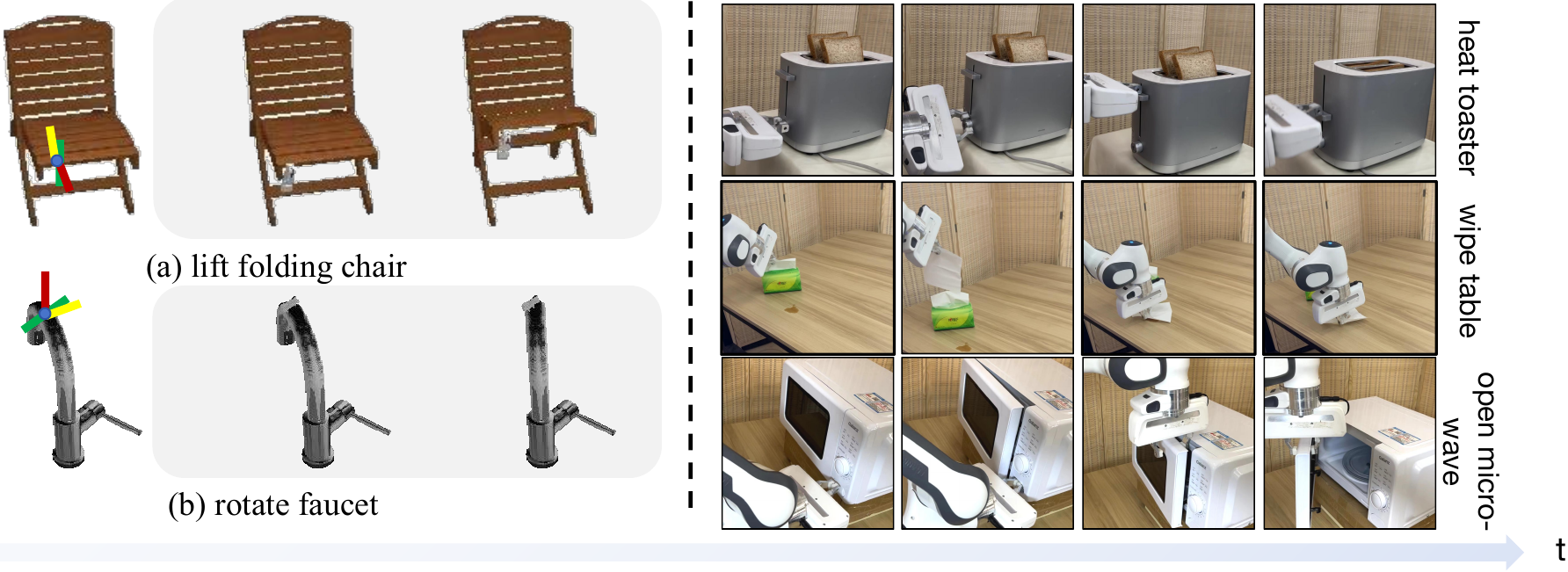}
   \caption{Visualization results in SAPIEN simulator and real world.}
   \label{fig:simviz}
\end{center}
% \vspace{-0.1cm}
\vspace{-0.5cm}
\end{figure*}
\begin{table}[ht]
% \vspace{-0.2cm}
\small
\centering
\begin{tabular}{c|cccc|c|c}
            \hline
           &  $a_{0}^{p}$ &  $a_{0}^{z}$&  $a_{0}^{y}$ &  $a_{0}^{m}$&Seen &Unseen \\\hline
           % Ex1&$\checkmark$&$\checkmark$&$\checkmark$&$\checkmark$ &0.16&0.10\\
           % Ex2 &$\checkmark$&$\checkmark$&$\checkmark$&$\checkmark$&0.69&0.68\\\hline
            Ex1 &$\checkmark$ &-&-&- &0.42 & 0.37 \\
    Ex2  & $\checkmark$ &$\checkmark$&-&-  &0.55&0.50  \\
    Ex3 & $\checkmark$ &$\checkmark$&$\checkmark$&-  &0.70&0.68\\\hline
            Ours &$\checkmark$&$\checkmark$& $\checkmark$ &$\checkmark$&0.74&0.72\\\hline
             Ours(w/o v.) &$\checkmark$&$\checkmark$& $\checkmark$ &$\checkmark$&0.69&0.68\\\hline
        \end{tabular}
    \caption{Ablation Study.}
        \label{tab:abla}

% \begin{minipage}[b]{0.4\linewidth}
% \centering
% \includegraphics[width=5cm, height=2.5cm]{./figure/seen-unseen.pdf}
% \captionof{figure}{Ablation on noise of input.}
% \label{fig:abla}
% \end{minipage}
\vspace{-0.5cm}
\end{table}
In Table \ref{tab:main}, both Ours(s) and Ours(f) demonstrate a substantial performance advantage over the baselines with manual generated prompts.
% , showcasing its proficiency in understanding input prompts and generating precise poses. 
For automatically generated prompts, the results are 0.64/0.62 on seen and unseen tasks, still outperforming the baselines.
It is important to note that by selecting the center of the bounding box generated by GroundingDiNO as the contact point input, there may be misalignment with the actual contact point, introducing noise into the positional prompt.
However, our results demonstrate the robustness of CrayonRobo in handling such input inaccuracies. 
This is because the model is trained to manipulate objects, it can, to some extent, correct the noise in the prompts. 
% Additionally, during training, we intentionally added noise to the data, further enabling the model to adapt and correct such discrepancies.
Simulator visualizations are shown in the left of Figure \ref{fig:simviz}, illustrating the prompt input, the robot's contact state with the object, and the final state after movement.
We further conduct experiments involving tasks with multiple steps, such as pulling a door and subsequently pushing it. Our method results in satisfactory performance: 0.69 and 0.68 on seen and unseen tasks, showing its potential in handling long-horizon tasks.

\subsection{Ablation Study}
% Ablation studies are presented in Table \ref{tab:sec}, with results achieved using a finger gripper as the actuator.

\subsubsection{Analysis on The Effect of Different Types of Prompt}
\label{sec: prompt type}

In Table~\ref{tab:abla} Ex1-Ex3, since our model is able to handle various input patterns thanks to the proposed training strategy,
we progressively introduce each type of prompt during testing.
When no $a_{0}^{m}$ is provided, we use rule-based approach~\citep{mo2021where2act} to determine the movement after contact.
Beginning with Ex1, where only a 2D position prompt is provided, the model achieves impressive performance with scores of 0.42/0.37. 
We compare our results with using AnyGrasp~\citep{fang2023anygrasp} to predict rotation given the same pixel coordinate, which results in lower scores of 0.35/0.31. 
This shows even without directional prompts, our model can accurately predict 3D poses, showcasing its ability to comprehend objects and predict appropriate poses based on positional prompts.
Moreover, comparing Ex2 and Ex3, we observe that by introducing more prompts during testing, the model achieves more accurate predictions. 

Additionally, to investigate the differential effects of visual and language prompts, in the last row of Table~\ref{tab:abla}, we enable the model to learn from images capturing only the object (without visual prompt), along with language containing instructional prompts. This configuration results in a performance drop compared to inputting prompts from both modalities. The reduced performance can be attributed to the difficulty the model has in effectively linking the given language prompt to the relevant objects. Consequently, it becomes challenging to make accurate object-centric manipulation predictions when relying solely on language prompts.
We thus conclude that prompts from both modalities work together to enhance the model's understanding and improve its pose prediction capabilities.
% \textit{Automatic generated prompt.}

% We can conclude with automated extract positional prompt, our model can still achieve promising results.  
\begin{figure}[ht]
\begin{center}
   \includegraphics[width=0.35\textwidth]{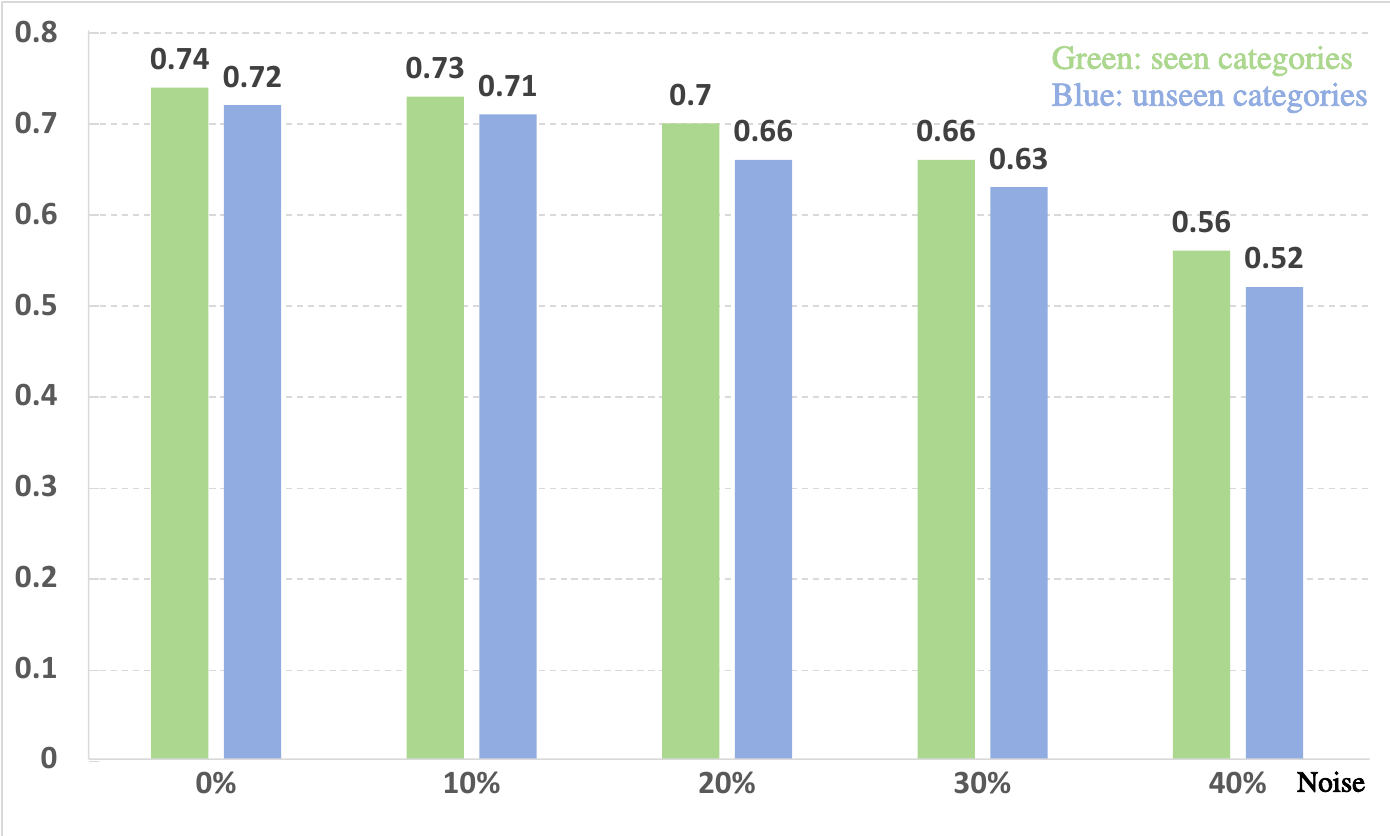}
   \caption{Robustness analysis on the noise in prompts.}
   \label{fig:noise}
\end{center}
% \vspace{-0.1cm}
% \vspace{-0.4cm}
\end{figure}

\subsubsection{Tolerance Analysis of Prompt Noise}
\label{sec: noise}
% We present the results using automatically generated prompts in Table. \ref{tab:main}. It is important to note that, because we select the center of the bounding box generated by GroundingDiNO as the contact point input, it may not perfectly align with the actual contact point, introducing some noise into the positional prompt. However, our results demonstrate the robustness of our approach in handling such input inaccuracies. Since the model is trained to manipulate objects, it can, to some extent, correct the noise in the prompts. Additionally, during training, we intentionally added noise to the data, further enabling the model to adapt and correct such discrepancies. Even with automatically generated prompts, our method achieves 0.64/0.62 on seen/unseen tasks, consistently outperforming baseline methods. 

% The ability of handling noise in positional prompt during testing is evaluated in Section \ref{sec:baseline}. 
We further introduce noise into the 2D directional prompt during testing while maintaining the accurate 2D contact point, in order to evaluate the model’s tolerance to input disturbances.
Random noise with a uniform distribution is added to the directional prompt, ranging from 10\%, 20\%, 30\%, to 40\% of the original directional value. The results, shown in Figure \ref{fig:noise}, indicate that with 10\% and 20\% noise, our method achieves performance levels comparable to those of the noise-free scenario. This demonstrates the model’s ability to handle noisy inputs during testing. However, with 30\% and 40\% noise, performance degradation occurs as the directional values deviate significantly from their intended targets—deviations that would be unlikely to happen if drawn by a human, as they appear visibly unreasonable. Despite this, the model is still capable of successfully executing tasks, even when the directional inputs are imperfect or provided by a non-expert. This highlights the model's resilience, not only to noise but also to imperfect or non-ideal inputs, allowing it to perform effectively even when the prompts are not perfectly accurate.

Ablation experiments regarding the effectiveness of each loss and failure case analysis are shown in Appendix.~\ref{sec:A5} and Appendix.~\ref{sec:A6}.
% \begin{figure*}[h]
% \begin{center}
%     % \includegraphics[scale=0.5, 
%    \includegraphics[width=0.98\textwidth]{./figure/vis_v7.pdf}
%    \caption{Illustration of executing the long-horizon tasks consisting of multiple atomic steps.}
%    \label{fig:viz}
% \end{center}
% \vspace{-0.6cm}
% \end{figure*}
\subsection{Real-world Experiment}
\label{sec:real}
\begin{table}
    \centering
        \small
        \begin{tabular}{c|ccccc}
        \toprule
        % \multicolumn{3}{c}{Seen Tasks} & \multicolumn{2}{c}{Unseen Tasks} \\
        % \cmidrule(lr){1-3} \cmidrule(lr){4-5}

        & \makecell{Open the\\trashcan} & \makecell{Open \\microwave*} & \makecell{Lift\\ lid}&\makecell{Wipe \\ table*}&\makecell{Heat \\ toaster*} \\
        \midrule
      dra.& 9/10  & 7/10&5/5  & 8/10&3/5 \\
       auto.& 8/10  & 6/10&5/5  & 8/10&2/5 \\
        w/o P.& -  & 5/10&- & 6/10&- \\
        \bottomrule
    \end{tabular}
        % \begin{tabular}{|c|c|c|c|c|}
        %     \hline
        %     \makecell{Lift \\ pan lid} & \makecell{Push \\ button} & \makecell{Slide \\ lever} & \makecell{Open micro- \\wave door}& \makecell{Open the\\trashcan} \\
        %     \hline
        %     5/5  & 3/5&3/5  & 4/5&5/5 \\
        %     \hline
        % \end{tabular}
        \captionof{table}{Real-world success rate. * contains multiple steps.}
        \label{tab:realexp}
        % \vspace{-0.5cm}
\end{table}
We conduct experiments involving interaction with various real-world objects without additional sim-to-real finetuning. 
Our setup includes a Franka Emika robotic arm equipped with its default parallel gripper, along with a RealSense 415 camera for capturing RGB images and depth maps.
% To address the sim-to-real problem effectively, we employ two key strategies:
% 1) During training, we leverage the LLaMA-Adapter pre-trained in the real world. 
% We employ a fine-tuning approach that focuses solely on updating adapters to enable the model to learn new downstream manipulation tasks. 
% This strategy allows the model to retain its robust perception abilities in the real world while acquiring novel skills. 
% 2) When collecting data in the simulator, we employ domain randomization to enhance scenario diversity.
% This involves varying elements such as object part poses and camera view angles, among others, to mitigate potential sim-to-real discrepancies.
In our real-world experiment, we employ the workstation's built-in image editor to directly draw crayon visual prompts on the images or automatic generating prompts, denoted as ``dra.'' and ``auto.'' in Table. \ref{tab:realexp} . We subsequently extract 2D vectors from the visual prompt, forming the text input. 
For each task, we select one or two shapes of objects and perform five trials each with a different camera view angle and object's initial pose. 
If the task has been completed, we consider the execution successful. 
To avoid a collision, we introduce a ``pre-move'' waypoint before reaching the predicted contact pose. This waypoint is positioned along the end-effector’s z-axis, at a distance away from the predicted contact pose.
Qualitative and quantitative results are presented in Table. \ref{tab:realexp} and the right part of Figure. \ref{fig:simviz}. 
The results demonstrate that CrayonRobo possesses strong zero-shot capability on novel tasks, as it interprets task execution prompts rather than relying solely on images or language processing, which can be challenging to handle in novel scenarios.
Meanwhile, it exhibits strong capability in long-horizon tasks, achieving a success rate comparable to that of single-step tasks.
% As shown in Table. \ref{tab:realexp}, the metric is the success rate. Visualizations are shown in the right part of Figure. \ref{fig:simviz}.

In the ``w/o P.'' row of Table. \ref{tab:realexp}, we further explore whether it is possible to fine-tune the model using prompts provided during the first set of executions for a specific task, and enable the model to perform subsequent executions independently, without requiring additional prompts. To investigate this, we selected two multi-step tasks: opening the microwave and wiping the table. The model is fine-tuned for 20 epochs using key frames and the corresponding predicted 3D poses from successful trials in the previous experiment as ground truth (GT), without providing any visual or text prompt inputs. During fine-tuning, the image input includes only the object, while the language input excludes all 2D prompts but incorporates the current robot state to ensure temporal consistency. The output format remains consistent with Figure \ref{fig:policy}. The fine-tuned model is then evaluated on these tasks, achieving success rates of 5/10 and 6/10, respectively. 
% More details of finetuning are in Appendix. 
\section{Conclusion}
We introduce a novel approach using comprehensive multi-modal prompts that explicitly convey both low-level actions and high-level planning in a simple manner.
We devise a training strategy that enables the model to comprehend the 2D visual-language prompt and predict accurate contact pose along with moving direction, ensuring the reliability of task execution. Such model-interpretable prompts show promising performance on both seen and unseen tasks, demonstrating its robustness. 
\textit{Limitation: }As for the limitation, though our method can not directly avoid obstacles, we can incorporate collision-free motion planner library like curobo~\citep{sundaralingam2023curobo} to realize this to some extent. 

\section*{Acknowledge}
This project was supported by National Youth Talent Support Program (8200800081) and National Natural Science Foundation of China (62376006).
% The method leverages a depth map that may be disturbed by reflective objects in the scene. We will further incorporate 

% While our proposed approach demonstrates generalization capabilities for novel manipulation tasks, several limitations remain to be addressed. 
% For example, the current visual prompt does not specify the dense trajectories, which requires additional caution in cases where obstacles are present along the trajectory.
% Therefore, future exploration could focus on using simple and straightforward visual prompts to convey more comprehensive information. 
% \clearpage
{
    \small
    \bibliographystyle{ieeenat_fullname}
    \bibliography{main}

\begin{thebibliography}{64}
\providecommand{\natexlab}[1]{#1}
\providecommand{\url}[1]{\texttt{#1}}
\expandafter\ifx\csname urlstyle\endcsname\relax
  \providecommand{\doi}[1]{doi: #1}\else
  \providecommand{\doi}{doi: \begingroup \urlstyle{rm}\Url}\fi

\bibitem[Achiam et~al.(2023)Achiam, Adler, Agarwal, Ahmad, Akkaya, Aleman, Almeida, Altenschmidt, Altman, Anadkat, et~al.]{achiam2023gpt}
Josh Achiam, Steven Adler, Sandhini Agarwal, Lama Ahmad, Ilge Akkaya, Florencia~Leoni Aleman, Diogo Almeida, Janko Altenschmidt, Sam Altman, Shyamal Anadkat, et~al.
\newblock Gpt-4 technical report.
\newblock \emph{arXiv preprint arXiv:2303.08774}, 2023.

\bibitem[Ahn et~al.(2022)Ahn, Brohan, Brown, Chebotar, Cortes, David, Finn, Fu, Gopalakrishnan, Hausman, et~al.]{ahn2022can}
Michael Ahn, Anthony Brohan, Noah Brown, Yevgen Chebotar, Omar Cortes, Byron David, Chelsea Finn, Chuyuan Fu, Keerthana Gopalakrishnan, Karol Hausman, et~al.
\newblock Do as i can, not as i say: Grounding language in robotic affordances.
\newblock \emph{arXiv preprint arXiv:2204.01691}, 2022.

\bibitem[An et~al.(2024)An, Geng, Chen, Li, Dou, and Dong]{an2024rgbmanip}
Boshi An, Yiran Geng, Kai Chen, Xiaoqi Li, Qi Dou, and Hao Dong.
\newblock Rgbmanip: Monocular image-based robotic manipulation through active object pose estimation.
\newblock In \emph{2024 IEEE International Conference on Robotics and Automation (ICRA)}, pages 7748--7755. IEEE, 2024.

\bibitem[Bahl et~al.(2023)Bahl, Mendonca, Chen, Jain, and Pathak]{bahl2023affordances}
Shikhar Bahl, Russell Mendonca, Lili Chen, Unnat Jain, and Deepak Pathak.
\newblock Affordances from human videos as a versatile representation for robotics.
\newblock In \emph{Proceedings of the IEEE/CVF Conference on Computer Vision and Pattern Recognition}, pages 13778--13790, 2023.

\bibitem[Belkhale et~al.(2024)Belkhale, Ding, Xiao, Sermanet, Vuong, Tompson, Chebotar, Dwibedi, and Sadigh]{belkhale2024rt}
Suneel Belkhale, Tianli Ding, Ted Xiao, Pierre Sermanet, Quon Vuong, Jonathan Tompson, Yevgen Chebotar, Debidatta Dwibedi, and Dorsa Sadigh.
\newblock Rt-h: Action hierarchies using language.
\newblock \emph{arXiv preprint arXiv:2403.01823}, 2024.

\bibitem[Black et~al.(2023)Black, Nakamoto, Atreya, Walke, Finn, Kumar, and Levine]{black2023zero}
Kevin Black, Mitsuhiko Nakamoto, Pranav Atreya, Homer Walke, Chelsea Finn, Aviral Kumar, and Sergey Levine.
\newblock Zero-shot robotic manipulation with pretrained image-editing diffusion models.
\newblock \emph{arXiv preprint arXiv:2310.10639}, 2023.

\bibitem[Bousmalis et~al.(2023)Bousmalis, Vezzani, Rao, Devin, Lee, Bauza, Davchev, Zhou, Gupta, Raju, et~al.]{bousmalis2023robocat}
Konstantinos Bousmalis, Giulia Vezzani, Dushyant Rao, Coline Devin, Alex~X Lee, Maria Bauza, Todor Davchev, Yuxiang Zhou, Agrim Gupta, Akhil Raju, et~al.
\newblock Robocat: A self-improving foundation agent for robotic manipulation.
\newblock \emph{arXiv preprint arXiv:2306.11706}, 2023.

\bibitem[Brohan et~al.(2022)Brohan, Brown, Carbajal, Chebotar, Dabis, Finn, Gopalakrishnan, Hausman, Herzog, Hsu, et~al.]{brohan2022rt}
Anthony Brohan, Noah Brown, Justice Carbajal, Yevgen Chebotar, Joseph Dabis, Chelsea Finn, Keerthana Gopalakrishnan, Karol Hausman, Alex Herzog, Jasmine Hsu, et~al.
\newblock Rt-1: Robotics transformer for real-world control at scale.
\newblock \emph{arXiv preprint arXiv:2212.06817}, 2022.

\bibitem[Brohan et~al.(2023)Brohan, Brown, Carbajal, Chebotar, Chen, Choromanski, Ding, Driess, Dubey, Finn, et~al.]{brohan2023rt}
Anthony Brohan, Noah Brown, Justice Carbajal, Yevgen Chebotar, Xi Chen, Krzysztof Choromanski, Tianli Ding, Danny Driess, Avinava Dubey, Chelsea Finn, et~al.
\newblock Rt-2: Vision-language-action models transfer web knowledge to robotic control.
\newblock \emph{arXiv preprint arXiv:2307.15818}, 2023.

\bibitem[Chane-Sane et~al.(2023)Chane-Sane, Schmid, and Laptev]{chane2023learning}
Elliot Chane-Sane, Cordelia Schmid, and Ivan Laptev.
\newblock Learning video-conditioned policies for unseen manipulation tasks.
\newblock In \emph{2023 IEEE International Conference on Robotics and Automation (ICRA)}, pages 909--916. IEEE, 2023.

\bibitem[Chi et~al.(2023)Chi, Feng, Du, Xu, Cousineau, Burchfiel, and Song]{chi2023diffusion}
Cheng Chi, Siyuan Feng, Yilun Du, Zhenjia Xu, Eric Cousineau, Benjamin Burchfiel, and Shuran Song.
\newblock Diffusion policy: Visuomotor policy learning via action diffusion.
\newblock \emph{arXiv preprint arXiv:2303.04137}, 2023.

\bibitem[Dai et~al.(2023)Dai, Pan, Sun, Ji, Xu, Liu, Wang, and Yang]{dai2023safe}
Josef Dai, Xuehai Pan, Ruiyang Sun, Jiaming Ji, Xinbo Xu, Mickel Liu, Yizhou Wang, and Yaodong Yang.
\newblock Safe rlhf: Safe reinforcement learning from human feedback.
\newblock \emph{arXiv preprint arXiv:2310.12773}, 2023.

\bibitem[Du et~al.(2023)Du, Yang, Florence, Xia, Wahid, Ichter, Sermanet, Yu, Abbeel, Tenenbaum, et~al.]{du2023video}
Yilun Du, Mengjiao Yang, Pete Florence, Fei Xia, Ayzaan Wahid, Brian Ichter, Pierre Sermanet, Tianhe Yu, Pieter Abbeel, Joshua~B Tenenbaum, et~al.
\newblock Video language planning.
\newblock \emph{arXiv preprint arXiv:2310.10625}, 2023.

\bibitem[Du et~al.(2024)Du, Yang, Dai, Dai, Nachum, Tenenbaum, Schuurmans, and Abbeel]{du2024learning}
Yilun Du, Sherry Yang, Bo Dai, Hanjun Dai, Ofir Nachum, Josh Tenenbaum, Dale Schuurmans, and Pieter Abbeel.
\newblock Learning universal policies via text-guided video generation.
\newblock \emph{Advances in Neural Information Processing Systems}, 36, 2024.

\bibitem[Eisner et~al.(2022)Eisner, Zhang, and Held]{eisner2022flowbot3d}
Ben Eisner, Harry Zhang, and David Held.
\newblock Flowbot3d: Learning 3d articulation flow to manipulate articulated objects.
\newblock \emph{arXiv preprint arXiv:2205.04382}, 2022.

\bibitem[Fang et~al.(2023)Fang, Wang, Fang, Gou, Liu, Yan, Liu, Xie, and Lu]{fang2023anygrasp}
Hao-Shu Fang, Chenxi Wang, Hongjie Fang, Minghao Gou, Jirong Liu, Hengxu Yan, Wenhai Liu, Yichen Xie, and Cewu Lu.
\newblock Anygrasp: Robust and efficient grasp perception in spatial and temporal domains.
\newblock \emph{IEEE Transactions on Robotics}, 2023.

\bibitem[Giraud et~al.(2005)Giraud, Langou, and Rozloznik]{giraud2005loss}
Luc Giraud, Julien Langou, and Miroslav Rozloznik.
\newblock The loss of orthogonality in the gram-schmidt orthogonalization process.
\newblock \emph{Computers \& Mathematics with Applications}, 50\penalty0 (7):\penalty0 1069--1075, 2005.

\bibitem[Goyal et~al.(2023)Goyal, Xu, Guo, Blukis, Chao, and Fox]{goyal2023rvt}
Ankit Goyal, Jie Xu, Yijie Guo, Valts Blukis, Yu-Wei Chao, and Dieter Fox.
\newblock Rvt: Robotic view transformer for 3d object manipulation.
\newblock In \emph{Conference on Robot Learning}, pages 694--710. PMLR, 2023.

\bibitem[Goyal et~al.(2024)Goyal, Blukis, Xu, Guo, Chao, and Fox]{goyal2024rvt}
Ankit Goyal, Valts Blukis, Jie Xu, Yijie Guo, Yu-Wei Chao, and Dieter Fox.
\newblock Rvt-2: Learning precise manipulation from few demonstrations.
\newblock \emph{arXiv preprint arXiv:2406.08545}, 2024.

\bibitem[Gu et~al.(2023)Gu, Kirmani, Wohlhart, Lu, Arenas, Rao, Yu, Fu, Gopalakrishnan, Xu, et~al.]{gu2023rt}
Jiayuan Gu, Sean Kirmani, Paul Wohlhart, Yao Lu, Montserrat~Gonzalez Arenas, Kanishka Rao, Wenhao Yu, Chuyuan Fu, Keerthana Gopalakrishnan, Zhuo Xu, et~al.
\newblock Rt-trajectory: Robotic task generalization via hindsight trajectory sketches.
\newblock \emph{arXiv preprint arXiv:2311.01977}, 2023.

\bibitem[Hu et~al.(2021)Hu, Shen, Wallis, Allen-Zhu, Li, Wang, Wang, and Chen]{hu2021lora}
Edward~J Hu, Yelong Shen, Phillip Wallis, Zeyuan Allen-Zhu, Yuanzhi Li, Shean Wang, Lu Wang, and Weizhu Chen.
\newblock Lora: Low-rank adaptation of large language models.
\newblock \emph{arXiv preprint arXiv:2106.09685}, 2021.

\bibitem[Huang et~al.(2024{\natexlab{a}})Huang, Chang, Liu, Zhu, Dong, Gao, Boularias, and Li]{huang2024a3vlm}
Siyuan Huang, Haonan Chang, Yuhan Liu, Yimeng Zhu, Hao Dong, Peng Gao, Abdeslam Boularias, and Hongsheng Li.
\newblock A3vlm: Actionable articulation-aware vision language model.
\newblock \emph{arXiv preprint arXiv:2406.07549}, 2024{\natexlab{a}}.

\bibitem[Huang et~al.(2024{\natexlab{b}})Huang, Ponomarenko, Jiang, Li, Hu, Gao, Li, and Dong]{huang2024manipvqa}
Siyuan Huang, Iaroslav Ponomarenko, Zhengkai Jiang, Xiaoqi Li, Xiaobin Hu, Peng Gao, Hongsheng Li, and Hao Dong.
\newblock Manipvqa: Injecting robotic affordance and physically grounded information into multi-modal large language models.
\newblock \emph{arXiv preprint arXiv:2403.11289}, 2024{\natexlab{b}}.

\bibitem[Huang et~al.(2023)Huang, Wang, Zhang, Li, Wu, and Fei-Fei]{huang2023voxposer}
Wenlong Huang, Chen Wang, Ruohan Zhang, Yunzhu Li, Jiajun Wu, and Li Fei-Fei.
\newblock Voxposer: Composable 3d value maps for robotic manipulation with language models.
\newblock \emph{arXiv preprint arXiv:2307.05973}, 2023.

\bibitem[Jia et~al.(2024)Jia, Liu, Chen, Gu, Wang, Luo, Lee, Wang, Wang, Zhang, et~al.]{jia2024lift3d}
Yueru Jia, Jiaming Liu, Sixiang Chen, Chenyang Gu, Zhilue Wang, Longzan Luo, Lily Lee, Pengwei Wang, Zhongyuan Wang, Renrui Zhang, et~al.
\newblock Lift3d foundation policy: Lifting 2d large-scale pretrained models for robust 3d robotic manipulation.
\newblock \emph{arXiv preprint arXiv:2411.18623}, 2024.

\bibitem[Jiang et~al.(2022)Jiang, Gupta, Zhang, Wang, Dou, Chen, Fei-Fei, Anandkumar, Zhu, and Fan]{jiang2022vima}
Yunfan Jiang, Agrim Gupta, Zichen Zhang, Guanzhi Wang, Yongqiang Dou, Yanjun Chen, Li Fei-Fei, Anima Anandkumar, Yuke Zhu, and Linxi Fan.
\newblock Vima: General robot manipulation with multimodal prompts.
\newblock \emph{arXiv preprint arXiv:2210.03094}, 2\penalty0 (3):\penalty0 6, 2022.

\bibitem[Johns(2021)]{johns2021coarse}
Edward Johns.
\newblock Coarse-to-fine imitation learning: Robot manipulation from a single demonstration.
\newblock In \emph{2021 IEEE international conference on robotics and automation (ICRA)}, pages 4613--4619. IEEE, 2021.

\bibitem[Ju et~al.(2024)Ju, Hu, Zhang, Zhang, Jiang, and Xu]{ju2024robo}
Yuanchen Ju, Kaizhe Hu, Guowei Zhang, Gu Zhang, Mingrun Jiang, and Huazhe Xu.
\newblock Robo-abc: Affordance generalization beyond categories via semantic correspondence for robot manipulation.
\newblock \emph{arXiv preprint arXiv:2401.07487}, 2024.

\bibitem[Ke et~al.(2024)Ke, Gkanatsios, and Fragkiadaki]{ke20243d}
Tsung-Wei Ke, Nikolaos Gkanatsios, and Katerina Fragkiadaki.
\newblock 3d diffuser actor: Policy diffusion with 3d scene representations.
\newblock \emph{arXiv preprint arXiv:2402.10885}, 2024.

\bibitem[Kim et~al.(2024)Kim, Pertsch, Karamcheti, Xiao, Balakrishna, Nair, Rafailov, Foster, Lam, Sanketi, et~al.]{kim2024openvla}
Moo~Jin Kim, Karl Pertsch, Siddharth Karamcheti, Ted Xiao, Ashwin Balakrishna, Suraj Nair, Rafael Rafailov, Ethan Foster, Grace Lam, Pannag Sanketi, et~al.
\newblock Openvla: An open-source vision-language-action model.
\newblock \emph{arXiv preprint arXiv:2406.09246}, 2024.

\bibitem[Li et~al.(2023)Li, Liu, Zhang, Yu, Xu, Wu, Cheang, Jing, Zhang, Liu, et~al.]{li2023vision}
Xinghang Li, Minghuan Liu, Hanbo Zhang, Cunjun Yu, Jie Xu, Hongtao Wu, Chilam Cheang, Ya Jing, Weinan Zhang, Huaping Liu, et~al.
\newblock Vision-language foundation models as effective robot imitators.
\newblock \emph{arXiv preprint arXiv:2311.01378}, 2023.

\bibitem[Li et~al.(2024)Li, Zhang, Geng, Geng, Long, Shen, Zhang, Liu, and Dong]{li2024manipllm}
Xiaoqi Li, Mingxu Zhang, Yiran Geng, Haoran Geng, Yuxing Long, Yan Shen, Renrui Zhang, Jiaming Liu, and Hao Dong.
\newblock Manipllm: Embodied multimodal large language model for object-centric robotic manipulation.
\newblock In \emph{Proceedings of the IEEE/CVF Conference on Computer Vision and Pattern Recognition}, pages 18061--18070, 2024.

\bibitem[Li et~al.(2025)Li, Deng, Zhang, Jang, Memme, Yu, Garrett, Ramos, Fox, Li, et~al.]{li2025hamster}
Yi Li, Yuquan Deng, Jesse Zhang, Joel Jang, Marius Memme, Raymond Yu, Caelan~Reed Garrett, Fabio Ramos, Dieter Fox, Anqi Li, et~al.
\newblock Hamster: Hierarchical action models for open-world robot manipulation.
\newblock \emph{arXiv preprint arXiv:2502.05485}, 2025.

\bibitem[Liang et~al.(2023)Liang, Huang, Xia, Xu, Hausman, Ichter, Florence, and Zeng]{liang2023code}
Jacky Liang, Wenlong Huang, Fei Xia, Peng Xu, Karol Hausman, Brian Ichter, Pete Florence, and Andy Zeng.
\newblock Code as policies: Language model programs for embodied control.
\newblock In \emph{2023 IEEE International Conference on Robotics and Automation (ICRA)}, pages 9493--9500. IEEE, 2023.

\bibitem[Liu et~al.(2024{\natexlab{a}})Liu, Fang, Abbeel, and Levine]{liu2024moka}
Fangchen Liu, Kuan Fang, Pieter Abbeel, and Sergey Levine.
\newblock Moka: Open-vocabulary robotic manipulation through mark-based visual prompting.
\newblock \emph{arXiv preprint arXiv:2403.03174}, 2024{\natexlab{a}}.

\bibitem[Liu et~al.(2024{\natexlab{b}})Liu, Liu, Wang, Lee, Zhou, An, Yang, Zhang, Guo, and Zhang]{liu2024robomamba}
Jiaming Liu, Mengzhen Liu, Zhenyu Wang, Lily Lee, Kaichen Zhou, Pengju An, Senqiao Yang, Renrui Zhang, Yandong Guo, and Shanghang Zhang.
\newblock Robomamba: Multimodal state space model for efficient robot reasoning and manipulation.
\newblock \emph{arXiv preprint arXiv:2406.04339}, 2024{\natexlab{b}}.

\bibitem[Liu et~al.(2025)Liu, Chen, An, Liu, Zhang, Gu, Li, Guo, Chen, Liu, et~al.]{liu2025hybridvla}
Jiaming Liu, Hao Chen, Pengju An, Zhuoyang Liu, Renrui Zhang, Chenyang Gu, Xiaoqi Li, Ziyu Guo, Sixiang Chen, Mengzhen Liu, et~al.
\newblock Hybridvla: Collaborative diffusion and autoregression in a unified vision-language-action model.
\newblock \emph{arXiv preprint arXiv:2503.10631}, 2025.

\bibitem[Liu et~al.(2023)Liu, Zeng, Ren, Li, Zhang, Yang, Li, Yang, Su, Zhu, et~al.]{liu2023grounding}
Shilong Liu, Zhaoyang Zeng, Tianhe Ren, Feng Li, Hao Zhang, Jie Yang, Chunyuan Li, Jianwei Yang, Hang Su, Jun Zhu, et~al.
\newblock Grounding dino: Marrying dino with grounded pre-training for open-set object detection.
\newblock \emph{arXiv preprint arXiv:2303.05499}, 2023.

\bibitem[Luo et~al.(2024)Luo, Hu, Xu, Tan, Berg, Sharma, Schaal, Finn, Gupta, and Levine]{luo2024serl}
Jianlan Luo, Zheyuan Hu, Charles Xu, You~Liang Tan, Jacob Berg, Archit Sharma, Stefan Schaal, Chelsea Finn, Abhishek Gupta, and Sergey Levine.
\newblock Serl: A software suite for sample-efficient robotic reinforcement learning.
\newblock \emph{arXiv preprint arXiv:2401.16013}, 2024.

\bibitem[Lynch and Sermanet(2020)]{lynch2020language}
Corey Lynch and Pierre Sermanet.
\newblock Language conditioned imitation learning over unstructured data.
\newblock \emph{arXiv preprint arXiv:2005.07648}, 2020.

\bibitem[Lynch et~al.(2020)Lynch, Khansari, Xiao, Kumar, Tompson, Levine, and Sermanet]{lynch2020learning}
Corey Lynch, Mohi Khansari, Ted Xiao, Vikash Kumar, Jonathan Tompson, Sergey Levine, and Pierre Sermanet.
\newblock Learning latent plans from play.
\newblock In \emph{Conference on robot learning}, pages 1113--1132. PMLR, 2020.

\bibitem[Mo et~al.(2021)Mo, Guibas, Mukadam, Gupta, and Tulsiani]{mo2021where2act}
Kaichun Mo, Leonidas~J Guibas, Mustafa Mukadam, Abhinav Gupta, and Shubham Tulsiani.
\newblock Where2act: From pixels to actions for articulated 3d objects.
\newblock In \emph{Proceedings of the IEEE/CVF International Conference on Computer Vision}, pages 6813--6823, 2021.

\bibitem[Nair et~al.(2022)Nair, Mitchell, Chen, Savarese, Finn, et~al.]{nair2022learning}
Suraj Nair, Eric Mitchell, Kevin Chen, Silvio Savarese, Chelsea Finn, et~al.
\newblock Learning language-conditioned robot behavior from offline data and crowd-sourced annotation.
\newblock In \emph{Conference on Robot Learning}, pages 1303--1315. PMLR, 2022.

\bibitem[Nasiriany et~al.(2024)Nasiriany, Xia, Yu, Xiao, Liang, Dasgupta, Xie, Driess, Wahid, Xu, et~al.]{nasiriany2024pivot}
Soroush Nasiriany, Fei Xia, Wenhao Yu, Ted Xiao, Jacky Liang, Ishita Dasgupta, Annie Xie, Danny Driess, Ayzaan Wahid, Zhuo Xu, et~al.
\newblock Pivot: Iterative visual prompting elicits actionable knowledge for vlms.
\newblock \emph{arXiv preprint arXiv:2402.07872}, 2024.

\bibitem[Nguyen and La(2019)]{nguyen2019review}
Hai Nguyen and Hung La.
\newblock Review of deep reinforcement learning for robot manipulation.
\newblock In \emph{2019 Third IEEE international conference on robotic computing (IRC)}, pages 590--595. IEEE, 2019.

\bibitem[Radford et~al.(2021)Radford, Kim, Hallacy, Ramesh, Goh, Agarwal, Sastry, Askell, Mishkin, Clark, et~al.]{radford2021learning}
Alec Radford, Jong~Wook Kim, Chris Hallacy, Aditya Ramesh, Gabriel Goh, Sandhini Agarwal, Girish Sastry, Amanda Askell, Pamela Mishkin, Jack Clark, et~al.
\newblock Learning transferable visual models from natural language supervision.
\newblock In \emph{International conference on machine learning}, pages 8748--8763. PMLR, 2021.

\bibitem[Shridhar et~al.(2022)Shridhar, Manuelli, and Fox]{shridhar2022cliport}
Mohit Shridhar, Lucas Manuelli, and Dieter Fox.
\newblock Cliport: What and where pathways for robotic manipulation.
\newblock In \emph{Conference on robot learning}, pages 894--906. PMLR, 2022.

\bibitem[Shridhar et~al.(2023)Shridhar, Manuelli, and Fox]{shridhar2023perceiver}
Mohit Shridhar, Lucas Manuelli, and Dieter Fox.
\newblock Perceiver-actor: A multi-task transformer for robotic manipulation.
\newblock In \emph{Conference on Robot Learning}, pages 785--799. PMLR, 2023.

\bibitem[Stone et~al.(2023)Stone, Xiao, Lu, Gopalakrishnan, Lee, Vuong, Wohlhart, Kirmani, Zitkovich, Xia, et~al.]{stone2023open}
Austin Stone, Ted Xiao, Yao Lu, Keerthana Gopalakrishnan, Kuang-Huei Lee, Quan Vuong, Paul Wohlhart, Sean Kirmani, Brianna Zitkovich, Fei Xia, et~al.
\newblock Open-world object manipulation using pre-trained vision-language models.
\newblock \emph{arXiv preprint arXiv:2303.00905}, 2023.

\bibitem[Sundaralingam et~al.(2023)Sundaralingam, Hari, Fishman, Garrett, Van~Wyk, Blukis, Millane, Oleynikova, Handa, Ramos, et~al.]{sundaralingam2023curobo}
Balakumar Sundaralingam, Siva Kumar~Sastry Hari, Adam Fishman, Caelan Garrett, Karl Van~Wyk, Valts Blukis, Alexander Millane, Helen Oleynikova, Ankur Handa, Fabio Ramos, et~al.
\newblock Curobo: Parallelized collision-free robot motion generation.
\newblock In \emph{2023 IEEE International Conference on Robotics and Automation (ICRA)}, pages 8112--8119. IEEE, 2023.

\bibitem[Sundaresan et~al.(2023)Sundaresan, Vuong, Gu, Xu, Xiao, Kirmani, Yu, Stark, Jain, Hausman, et~al.]{sundaresan2023rt}
Priya Sundaresan, Quan Vuong, Jiayuan Gu, Peng Xu, Ted Xiao, Sean Kirmani, Tianhe Yu, Michael Stark, Ajinkya Jain, Karol Hausman, et~al.
\newblock Rt-sketch: Goal-conditioned imitation learning from hand-drawn sketches.
\newblock 2023.

\bibitem[Team et~al.(2023)Team, Anil, Borgeaud, Wu, Alayrac, Yu, Soricut, Schalkwyk, Dai, Hauth, et~al.]{team2023gemini}
Gemini Team, Rohan Anil, Sebastian Borgeaud, Yonghui Wu, Jean-Baptiste Alayrac, Jiahui Yu, Radu Soricut, Johan Schalkwyk, Andrew~M Dai, Anja Hauth, et~al.
\newblock Gemini: a family of highly capable multimodal models.
\newblock \emph{arXiv preprint arXiv:2312.11805}, 2023.

\bibitem[Touvron et~al.(2023)Touvron, Lavril, Izacard, Martinet, Lachaux, Lacroix, Rozi{\`e}re, Goyal, Hambro, Azhar, et~al.]{touvron2023llama}
Hugo Touvron, Thibaut Lavril, Gautier Izacard, Xavier Martinet, Marie-Anne Lachaux, Timoth{\'e}e Lacroix, Baptiste Rozi{\`e}re, Naman Goyal, Eric Hambro, Faisal Azhar, et~al.
\newblock Llama: Open and efficient foundation language models.
\newblock \emph{arXiv preprint arXiv:2302.13971}, 2023.

\bibitem[Wen et~al.(2023)Wen, Lin, So, Chen, Dou, Gao, and Abbeel]{wen2023any}
Chuan Wen, Xingyu Lin, John So, Kai Chen, Qi Dou, Yang Gao, and Pieter Abbeel.
\newblock Any-point trajectory modeling for policy learning.
\newblock \emph{arXiv preprint arXiv:2401.00025}, 2023.

\bibitem[Xiang et~al.(2020)Xiang, Qin, Mo, Xia, Zhu, Liu, Liu, Jiang, Yuan, Wang, Yi, Chang, Guibas, and Su]{Xiang_2020_SAPIEN}
Fanbo Xiang, Yuzhe Qin, Kaichun Mo, Yikuan Xia, Hao Zhu, Fangchen Liu, Minghua Liu, Hanxiao Jiang, Yifu Yuan, He Wang, Li Yi, Angel~X. Chang, Leonidas~J. Guibas, and Hao Su.
\newblock {SAPIEN}: A simulated part-based interactive environment.
\newblock In \emph{The IEEE Conference on Computer Vision and Pattern Recognition (CVPR)}, 2020.

\bibitem[Xiao et~al.(2022)Xiao, Chan, Sermanet, Wahid, Brohan, Hausman, Levine, and Tompson]{xiao2022robotic}
Ted Xiao, Harris Chan, Pierre Sermanet, Ayzaan Wahid, Anthony Brohan, Karol Hausman, Sergey Levine, and Jonathan Tompson.
\newblock Robotic skill acquisition via instruction augmentation with vision-language models.
\newblock \emph{arXiv preprint arXiv:2211.11736}, 2022.

\bibitem[Xiong et~al.(2024)Xiong, Shen, Li, Zhou, Liu, Wang, and Dong]{xiong2024aic}
Chuyan Xiong, Chengyu Shen, Xiaoqi Li, Kaichen Zhou, Jiaming Liu, Ruiping Wang, and Hao Dong.
\newblock Aic mllm: Autonomous interactive correction mllm for robust robotic manipulation.
\newblock \emph{arXiv preprint arXiv:2406.11548}, 2024.

\bibitem[Xu et~al.(2022)Xu, He, and Song]{xu2022universal}
Zhenjia Xu, Zhanpeng He, and Shuran Song.
\newblock Universal manipulation policy network for articulated objects.
\newblock \emph{IEEE Robotics and Automation Letters}, 7\penalty0 (2):\penalty0 2447--2454, 2022.

\bibitem[Yang et~al.(2023{\natexlab{a}})Yang, Tan, Jin, Liu, Fu, Song, and Wang]{yang2023pave}
Jiange Yang, Wenhui Tan, Chuhao Jin, Bei Liu, Jianlong Fu, Ruihua Song, and Limin Wang.
\newblock Pave the way to grasp anything: Transferring foundation models for universal pick-place robots.
\newblock \emph{arXiv preprint arXiv:2306.05716}, 2023{\natexlab{a}}.

\bibitem[Yang et~al.(2023{\natexlab{b}})Yang, Du, Ghasemipour, Tompson, Schuurmans, and Abbeel]{yang2023learning}
Mengjiao Yang, Yilun Du, Kamyar Ghasemipour, Jonathan Tompson, Dale Schuurmans, and Pieter Abbeel.
\newblock Learning interactive real-world simulators.
\newblock \emph{arXiv preprint arXiv:2310.06114}, 2023{\natexlab{b}}.

\bibitem[Yuan et~al.(2023)Yuan, Murali, Mousavian, and Fox]{yuan2023m2t2}
Wentao Yuan, Adithyavairavan Murali, Arsalan Mousavian, and Dieter Fox.
\newblock M2t2: Multi-task masked transformer for object-centric pick and place.
\newblock In \emph{7th Annual Conference on Robot Learning}, 2023.

\bibitem[Yuan et~al.(2024)Yuan, Duan, Blukis, Pumacay, Krishna, Murali, Mousavian, and Fox]{yuan2024robopoint}
Wentao Yuan, Jiafei Duan, Valts Blukis, Wilbert Pumacay, Ranjay Krishna, Adithyavairavan Murali, Arsalan Mousavian, and Dieter Fox.
\newblock Robopoint: A vision-language model for spatial affordance prediction for robotics.
\newblock \emph{arXiv preprint arXiv:2406.10721}, 2024.

\bibitem[Ze et~al.(2024)Ze, Zhang, Zhang, Hu, Wang, and Xu]{ze20243d}
Yanjie Ze, Gu Zhang, Kangning Zhang, Chenyuan Hu, Muhan Wang, and Huazhe Xu.
\newblock 3d diffusion policy.
\newblock \emph{arXiv preprint arXiv:2403.03954}, 2024.

\bibitem[Zhong et~al.(2023)Zhong, Zheng, Zheng, Zhao, Yi, Mu, Wang, Li, Zhou, Yang, et~al.]{zhong20233d}
Chengliang Zhong, Yuhang Zheng, Yupeng Zheng, Hao Zhao, Li Yi, Xiaodong Mu, Ling Wang, Pengfei Li, Guyue Zhou, Chao Yang, et~al.
\newblock 3d implicit transporter for temporally consistent keypoint discovery.
\newblock In \emph{Proceedings of the IEEE/CVF International Conference on Computer Vision}, pages 3869--3880, 2023.

\end{thebibliography}
}
\appendix
\section{Input Details}
\label{sec:A1}

\begin{figure*}[ht]
\begin{center}
   \includegraphics[width=0.75\textwidth]{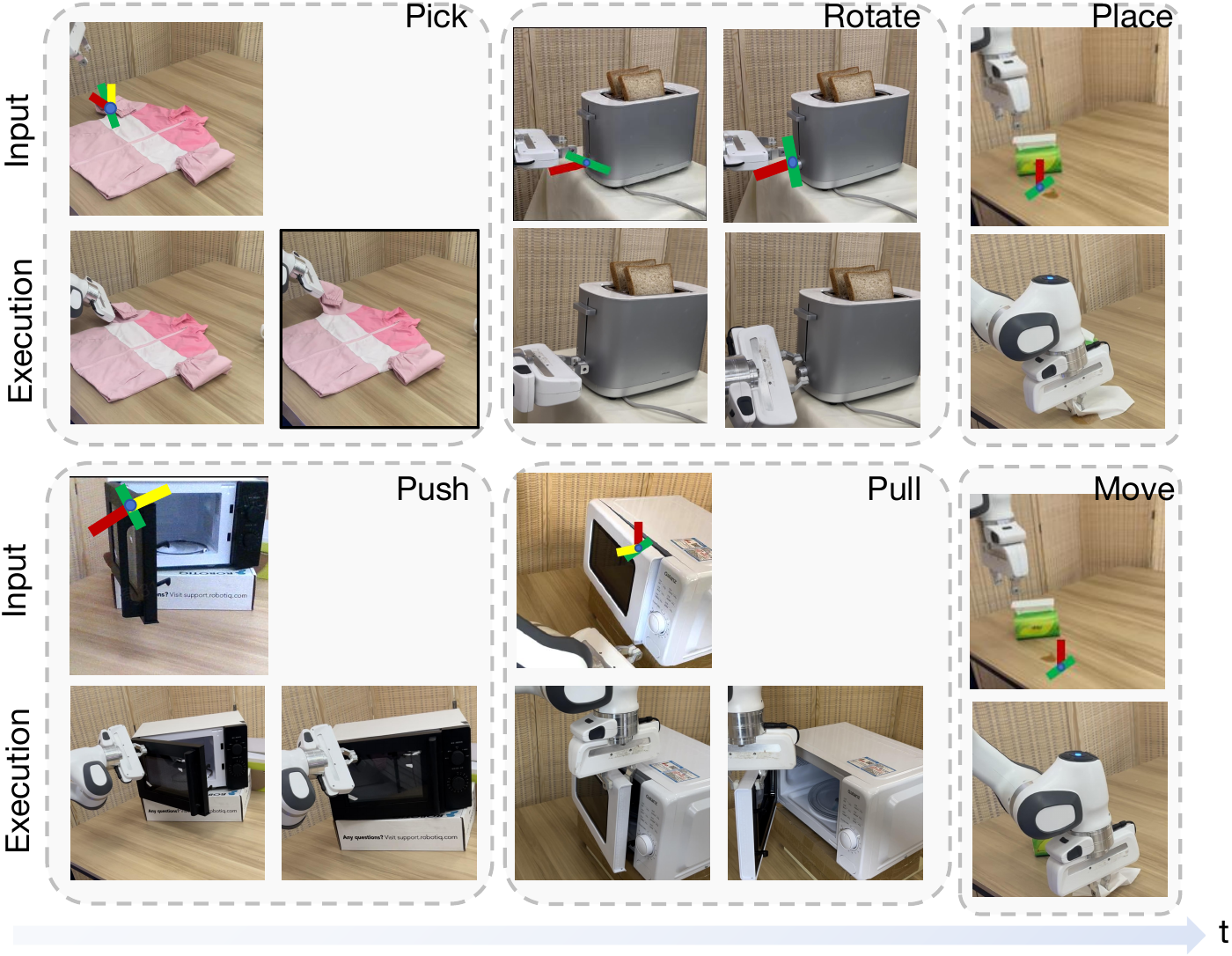}

\end{center}
% \vspace{-0.1cm}
\vspace{-0.2cm}
\caption{The completion of tyical key-frame tasks.}
\label{fig:atomic}
\end{figure*}
We outline the complete input language and ground-truth supervision in Section3.3.2:

Input(a): ``Predict the contact point and orientation for manipulating the object. The hints in the image include the contact point with a blue dot. Specifically, the contact point is at $a_{0}^{p}$."

Input(b): ``Predict the contact point and orientation for manipulating the object. The hints in the image include a blue dot for the contact point and a red line for the gripper z-axis 2D direction. Specifically, the contact point is at $a_{0}^{p}$, and the gripper z-axis 2D direction is $a_{0}^{z}$."

Input(c): ``Predict the contact point and orientation for manipulating the object. The hints in the image include a blue dot for the contact point, a red line for the gripper z-axis 2D direction, and a green line for the gripper y-axis 2D direction. Specifically, the contact point is at $a_{0}^{p}$, the gripper z-axis 2D direction is $a_{0}^{z}$, and the gripper y-axis 2D direction is $a_{0}^{y}$."

Input(d): ``Predict the contact point and orientation for manipulating the object. The hints in the image include a blue dot for the contact point, a red line for the gripper z-axis 2D direction, a green line for the gripper y-axis 2D direction, and a yellow line for the moving 2D direction. Specifically, the contact point is at $a_{0}^{p}$, the gripper z-axis 2D direction is $a_{0}^{z}$, the gripper y-axis 2D direction is $a_{0}^{y}$, and the gripper moving 2D direction is $a_{0}^{m}$."

Ground truth: ``The contact point is at $a_{0}^{p}$, the gripper z-axis 3D direction is $a_{0}^{Z'}$, the gripper y-axis 3D direction is $a_{0}^{Y'}$, and the moving 3D direction is $a_{0}^{M'}$."

\section{Data Collection Details}
\label{sec:A2}

The size of the training dataset is around 10,000. Regarding the variation between training and testing data, the specific variations can be divided into two aspects: 1) asset variation and 2) camera view variation.

\textit{Asset Variation}: We use 20 categories from PartNet-mobility~\cite{Xiang_2020_SAPIEN} for seen objects and reserve the remaining 10 categories for unseen objects to analyze whether CrayonRobo can generalize to novel categories. Specifically, we further divide the seen objects into 1,037 training shapes and 489 testing shapes, using only the training shapes to construct the training data. Thus, the shapes of the seen objects encountered during training and testing are different. For unseen categories, there are a total of 274 shapes, which are used exclusively in the testing data.

\textit{Camera View Variation}: We observe the object in the scene from an RGB-D camera with known intrinsics, mounted 4.5 to 5.5 units away from the object, facing its center. The camera is located in the upper hemisphere of the object with a random azimuth between \ang{45} and \ang{-45}, and a random altitude between \ang{30} and \ang{60}. We initialize the starting pose for each articulated part randomly between its rest joint state (fully closed) and any position up to half of its joint state (half-opened). In both the training and testing phases, the object is placed and captured randomly within the aforementioned scope.

\section{Details of RT-Trajectory Replication}
% \subsection{}
\label{sec:A3}
Since the code for RT-Trajectory is not publicly available, we replicate its method based on the paper's description. During data collection in the simulator, we record the 3D position of the end-effector and project it onto the camera frame to create the corresponding 2D trajectory. Given that the tasks are atomic, consisting of a single step (e.g., opening a door), color grading is unnecessary. Instead, we mark the start and end positions, as well as the gripper state, by drawing blue and green circles, respectively.

We use the same backbone as in our model, the LLaMA-adapter, and fine-tune it to process both the trajectory image and the current object image. This allows the model to output the 6DoF poses required to complete the tasks. The same training and testing splits are applied, resulting in an average success rate of 0.57 on seen categories and 0.52 on unseen categories for RT-Trajectory, while our model achieves 0.74 and 0.72, respectively.

Further investigation reveals that in our replication of RT-Trajectory, while the method accurately captures the end-effector's trajectory position, the rotation estimation is not precise enough for interacting with articulated objects. Unlike tasks such as pick-and-place, where the end-effector's rotation is relatively uniform, interactions with articulated objects demand more diverse and complex rotational adjustments, making it challenging for RT-Trajectory to learn effectively. This also highlights the need to provide directional prompts for the model to interpret.

\section{Visual Demonstration of Typical Atomic Tasks}
\label{sec:A4}
In Figure. \ref{fig:atomic}, we demonstrate how the key-frame tasks discussed in Section 3.4.1 are executed. We focus on the ``rotation button" task, which is unique because it does not involve the translation of the end effector but rather the rotation of the last joint. Using visual prompts from two key frames, we predict two poses. By analyzing the predicted poses, we observe that the only difference between them is the orientation of the end effector along the y-axis. We then pass these predictions to the ROS package for inverse kinematics (IK) computation, which allows us to achieve the desired rotation effect at the last joint.

\section{The Effectiveness of Each Loss.} 
\label{sec:A5}
\begin{table}[h]
% \vspace{-0.4cm}   
\begin{center}
\small
\begin{tabular}{c|ccc|cc} % Adjusted column width to match the content
    \hline
      & $\mathcal{L}_{T}$ & $\mathcal{L}_{O}$ & $\mathcal{L}_{P}$ & Seen & Unseen \\
    \hline
    Ex1 & $\checkmark$ & - & - & 0.68 & 0.57 \\
    Ex2 & $\checkmark$ & $\checkmark$ & - & 0.71 & 0.70 \\
    \rowcolor{verylightgrey} Ours & $\checkmark$ & $\checkmark$ & $\checkmark$ & 0.74 & 0.72 \\
    \hline
\end{tabular}

\end{center}
\caption{The effectiveness of each loss}
\label{tab:loss}

\end{table}

In Table. \ref{tab:loss}, we progressively introduce each loss objective during training: Ex1 involves training solely with $\mathcal{L}_{T}$; Ex2 combines $\mathcal{L}_{T}$ with $\mathcal{L}_{O}$; and Ours integrates $\mathcal{L}_{T}$ with both $\mathcal{L}_{O}$ and $\mathcal{L}_{P}$.
Comparing Ex2 and Ex1, we observe that incorporating $\mathcal{L}_{O}$ enhances accuracy by explicitly enforcing the orthogonality constraints between the z-axis and y-axis directions. 
Furthermore, the addition of $\mathcal{L}_{P}$ in Ours results in a further accuracy improvement compared to Ex2, showing its effectiveness in capturing the correlation between 2D prompts and 3D directions.

\section{Failure Analysis} 
\label{sec:A6}
We analyze the failure cases in real-world: for the push button step in the open microwave task, the excessive reactive force during button pressing prevented the robotic arm from completing the push successfully. For the slide lever step in the heat toaster task, the gripper fingers we use are too short, which sometimes prevent them from firmly contacting the lever during movement.

\end{document}